\documentclass{fairmeta}

\input{glyphtounicode}
\microtypesetup{expansion=false}
\usepackage{amsmath,amssymb,mathtools}
\usepackage{array}
\usepackage{colortbl}
\usepackage{float}
\usepackage{algpseudocode}
\usepackage{longtable}
\usetikzlibrary{positioning,arrows.meta,fit,calc,backgrounds}
\floatstyle{ruled}
\newfloat{algorithm}{tbp}{loa}
\floatname{algorithm}{Algorithm}
\crefname{algorithm}{algorithm}{algorithms}
\Crefname{algorithm}{Algorithm}{Algorithms}
\definecolor{metateal}{HTML}{008A78}
\definecolor{metapurple}{HTML}{7B61A8}
\definecolor{metagold}{HTML}{E89A20}

\ifnum 1 > 0
\newcommand{\hamed}[1]{{\color{red}[Hamed: #1]}}
\newcommand{\zs}[1]{{\color{cyan}[Zhouxing: #1]}}
\newcommand{\rui}[1]{{\color{orange}[Rui: #1]}}
\newcommand{\shangjian}[1]{{\color{metablue}[Shangjian: #1]}}
\else
\newcommand{\hamed}[1]{}
\newcommand{\zs}[1]{}
\newcommand{\rui}[1]{}
\newcommand{\shangjian}[1]{}
\fi

\newcommand{\dopsd}{\textsf{Dynamic Co-Evolution}}

\newcommand{\sg}{\operatorname{stopgrad}}
\newcommand{\eos}{\ensuremath{\mathrm{EOS}}}

\newcommand{\resulttok}[1]{\textcolor{metafg!90}{#1}}
\newcommand{\bestacc}[1]{\textcolor{metablue}{\bfseries #1}}

\title{Recursive Self-Improvement via On-Policy Distillation for Reasoning}

\author[1,2,*]{Shangjian Yin}
\author[1]{Zehao Zhao}
\author[1]{Kavosh Asadi}
\author[1]{Rui Liu}
\author[1]{Yuchen Lu}
\author[1]{Shike Mei}
\author[1]{Hang Cui}
\author[1]{Luke Simon}
\author[2]{Zhouxing Shi}
\author[1]{Hamed Firooz}

\affiliation[1]{Meta AI}
\affiliation[2]{University of California, Riverside}
\contribution[*]{Work done at Meta}

\abstract{On-policy distillation (OPD) trains a student model by having it generate trajectories, then matching its next-token predictions with an external teacher's next-token predictions. This provides a dense, token-level supervision to the student. On-policy
self-distillation (OPSD) eliminates the need for the external teacher. Specifically, a second frozen copy of the student model, now given the ground truth in its context, serves as the teacher. The student model only receives the problem and learns to mimic the privileged teacher model, while the teacher remains frozen throughout training. Previous work showed that freezing the teacher is useful for training stability, but we argue that this can prevent the teacher from incorporating the improvements learned by the student during training.
Our primary contribution is to address this limitation with a recursive framework built around two complementary components.
First, we let the privileged teacher to co-evolve with the student so that revision learned in
one round can guide the next, a process we refer to as \textbf{Dynamic Co-Evolution (DCE)}. Second,
because stronger revision can also make responses too verbose and self-critical, we additionally train on
shorter, verified rewrites of the model's own on-policy responses. We call this
complementary objective \textbf{Self-Refined Concise
Learning (SRCL)}.
Overall, our comprehensive evaluations show that DCE+SRCL outperforms OPSD across multiple model
scales and four competition-level mathematics benchmarks. Specifically, on Qwen3-8B, DCE+SRCL reaches 65.97\% Average@12,
outperforming OPSD by 35.62 percentage points while reducing mean output length by 7.80\% relative
to DCE alone.}

\date{September 11, 2026}

\begin{document}

\maketitle

\section{Introduction}
\label{sec:introduction}

Post-training has played a central role in recent advances in LLM reasoning. A prominent approach is
reinforcement learning with verifiable rewards (RLVR), which uses an automatic checker to assign an
outcome reward to a completed solution~\citep{shao2024deepseekmath,yu2025dapo}. Because correctness
can be verified without annotating every intermediate step, RLVR scales without step-level
supervision and can elicit longer derivations, intermediate verification, backtracking, and
self-correction~\citep{guo2025deepseek}. However, outcome rewards provide only coarse,
sequence-level supervision: they indicate whether the final answer is correct, but not where the
reasoning went wrong or how it should be revised.
Process supervision provides more local feedback, but generally requires
step-level annotations or a separately trained verifier~\citep{lightman2023let,zhang2025lessons}.

Between sparse outcome rewards and costly process supervision, on-policy distillation (OPD) offers a
third route: dense token-level targets on trajectories sampled from the model being trained
~\citep{agarwal2024policy,lu2025onpolicydistillation}. This model serves as the student, while a
separate, stronger teacher provides next-token logits for each generated prefix. These logits provide a richer
token-level training signal than the student's own predictions. However, standard OPD still
requires a capable external teacher, whose guidance is not conditioned on a verified solution for
the current problem. The student is therefore trained to match the teacher's token-level
predictions, potentially inheriting its errors as well as its strengths.
On-policy self-distillation (OPSD)~\citep{zhao2026selfdistilled} addresses these limitations using two roles initialized from
the same base checkpoint. During training, the student model is given only the problem and
generates an on-policy response. 
A copy of the base checkpoint is additionally given the ground-truth solution and serves as the privileged teacher. 
At each prefix of the student's response, the teacher
receives the problem, the ground-truth solution, and the same generated prefix, then supplies next-token
logits.
OPSD minimizes the divergence between the student and teacher predictions, transferring
gold-conditioned guidance to the student. %
To stabilize training, the teacher remains frozen in \citet{zhao2026selfdistilled}.

\begin{figure}[!t]
    \centering
    \includegraphics[width=0.98\textwidth]{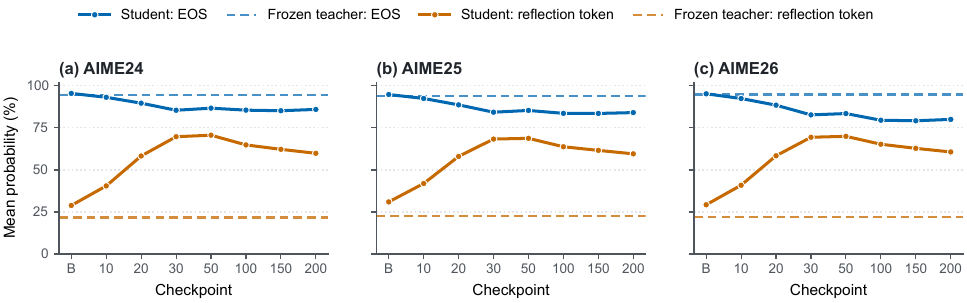}
    \caption{\textbf{Frozen teacher favors termination over reflection.} On fixed incorrect
    trajectories $Y^-$, the evolving student (solid) and gold-conditioned frozen teacher (dashed)
    score identical prefixes. Blue: $p(\eos\mid Y^-)$; orange:
    $p(r_t\mid Y^-_{<t})$ at observed reflection positions $t$.}
    \label{fig:frozen-teacher-probe}
\end{figure}

Does access to the ground-truth solution provide informative guidance at every student-generated
prefix? We probe this question on fixed incorrect Qwen3-8B trajectories from AIME 2024, AIME 2025,
and AIME 2026~\citep{maaAIME}, comparing evolving student checkpoints with the frozen,
gold-conditioned teacher on the same generated prefixes. Despite seeing the verified solution, the
frozen teacher assigns an average probability of 94.5\% to EOS at incorrect response endpoints but
only 22.0\% to observed reflection cues such as \texttt{Wait}, a common reflection marker in
reasoning models~\citep{wang2025wait} (\cref{fig:frozen-teacher-probe}), indicating limited guidance from the teacher on how to recover from incorrect reasoning. Meanwhile, the student
becomes substantially more likely to reflect during training, creating a growing mismatch between
the evolving model and its fixed supervisor. 

This mismatch exposes a limitation of freezing the privileged teacher. Revision behaviors such as
self-verification, backtracking, and error correction can strengthen during post-training
~\citep{guo2025deepseek,zhu2025emergence}, but a teacher fixed at the initial checkpoint cannot
acquire these emerging capabilities. Thus, even when conditioned on the verified solution, its
guidance may become increasingly misaligned with the student trajectories encountered later in
training. This observation motivates a privileged teacher that
evolves with the student.

We therefore introduce \dopsd\ (DCE), in which the privileged teacher evolves alongside the student. After each update, the resulting checkpoint initializes both the next student and a detached, gold-conditioned privileged teacher. Revision behavior acquired in one round can therefore improve the supervision provided in the next, creating a recursive self-improvement process. However, strengthening revision introduces a second challenge. More frequent checking,
and backtracking
can improve recovery from mistakes while also making reasoning unnecessarily long. We therefore pair DCE with \textbf{Self-Refined Concise Learning (SRCL)}, which trains on shorter, answer-verified rewrites of the same on-policy responses. DCE improves the model's ability to revise its reasoning, while SRCL encourages it to retain that capability without unnecessary token cost.

Our contributions can be summarized as follows:
\begin{itemize}
    \setlength{\itemsep}{2pt}

    \item We introduce \textbf{Dynamic Co-Evolution (DCE)}, a recursive on-policy
    self-distillation framework in which each updated checkpoint initializes both the next student
    and a detached, gold-conditioned privileged teacher. Across four competition-level mathematical
    reasoning benchmarks, DCE improves Average@12 accuracy over OPSD by 35.41, 37.15, and
    12.98 percentage points at 8B, 4B, and 1.7B, respectively.

    \item We introduce \textbf{Self-Refined Concise Learning (SRCL)} to control the reasoning cost
    that can accompany stronger revision. SRCL trains on accepted, shorter self-refinements of the
    same on-policy responses. Under our lowest tested 8K evaluation budget, DCE+SRCL achieves
    35.07\% Average@12 accuracy with 7,500 mean output tokens per response, outperforming the
    OPSD test-time-scaling control by 6.60 percentage points while generating approximately
    692 fewer tokens per response.

    \item We show that the privileged teacher learns to provide stronger revision guidance as it
    evolves with the student. Matched-budget controls show that longer generation alone does not
    explain the gains. Updating the teacher every round also outperforms frozen, EMA, and periodic
    alternatives. Fixed-trace probes corroborate the teacher's improved revision behavior across
    three AIME cohorts. Its endpoint EOS probability falls from 90.4\% to 41.3\%, while its
    probability on observed reflection tokens rises from 32.8\% to 77.4\%.
\end{itemize}

\section{Related Work}
\label{sec:related-work}

\subsection{Reflection and Verification in Reasoning Post-Training}

Outcome-supervised post-training can elicit reflection-like behaviors without step-level labels.
GRPO removes the learned critic used by PPO and estimates advantages from relative rewards within
a sampled group, making outcome-based training practical for mathematical
reasoning~\citep{shao2024deepseekmath}. DAPO introduces clip-higher, dynamic
sampling, and token-level policy loss to improve the stability and efficiency of this recipe at
scale~\citep{yu2025dapo}. DeepSeek-R1 further demonstrates that outcome-based training alone can
produce rechecking, backtracking, and ``aha moments'' without explicit reflection
supervision~\citep{guo2025deepseek}. Parallel efforts extend RL-based post-training beyond closed-form
math to policy-grounded content moderation~\citep{firooz2025scaling}, open-ended environments
requiring generalization without fixed-answer verification~\citep{yin2026grlo}, and non-verifiable
tasks balancing objective reasoning gains with subjective
alignment~\citep{yin-shi-2026-individual}. Despite this progress, outcome-level rewards score
completed responses as a whole and do not supervise
the local transition where the model identifies and repairs a specific error.

A closer look suggests that these reflection-like behaviors may not be newly learned through
training. R1-Zero reproductions find similar behaviors already present in some base models and
attribute part of the GRPO effect to a length bias toward longer, often incorrect,
responses~\citep{liu2025understanding}. Across ten base models, response length and verification
behavior do not reliably emerge together~\citep{zeng2025simplerl}. Activation-space analysis
further reveals a latent, though rare, capacity for reflection that exists before any
RLVR~\citep{zhu2025emergence}. Together, these results suggest that outcome-based post-training
amplifies a pre-existing behavioral prior rather than teaching the model how to recover from a
particular wrong prefix.

Even when reflection does appear in generation, it does not reliably correct errors. Without
external feedback, prompting a model to revise its own answer can reduce
accuracy~\citep{huang2023cannot}, and hidden-state probes reveal correctness signals that the
model's generation does not always exploit~\citep{zhang2025reasoningright,lee2025geometry}. The
gap between latent awareness and effective revision motivates a different form of supervision:
DCE provides dense, token-level guidance on the model's own incorrect reasoning, directly
training the transition from error recognition to successful revision.

\subsection{On-Policy and Privileged Self-Distillation}

OPD trains on responses sampled from the model being optimized, while a teacher supplies
next-token supervision along those same responses
~\citep{agarwal2024policy,lu2025onpolicydistillation}. OPSD removes the external teacher by assigning
a frozen copy of the initial checkpoint to the privileged role: the student sees only the problem,
whereas the teacher also receives a verified solution~\citep{zhao2026selfdistilled}. Follow-up
methods such as RLSD and RLCSD combine this signal with reinforcement learning
~\citep{yang2026rlsd,pan2026rlcsd}.

Formally, let $p_\theta$ denote the language model with parameters $\theta$. The student conditions
on a problem $x$ and generates a response $y\sim p_\theta(\cdot\mid x)$. The privileged teacher
scores each prefix $y_{<t}$ after additionally receiving a verified solution $g$ and a transition
instruction $\tau$, forming the privileged context
$\mathcal C_t^{\mathrm{ref}}(x,g,\tau,y_{<t})$. Let $q$ and $p$ denote the resulting teacher and
student next-token distributions, respectively, and let $\Delta(q,p)$ denote a divergence between
them. Our main experiments use Forward KL,
$\Delta(q,p)=D_{\mathrm{KL}}(q\|p)=\sum_v q(v)\log\frac{q(v)}{p(v)}$; alternatives are compared
in \cref{sec:divergence-control}. Standard OPSD minimizes
\begin{equation}
 \mathcal{L}_{\mathrm{fixed}}(\theta)
 =\mathbb{E}_{(x,g),y}\!\left[
   \frac{1}{|y|}\sum_{t=1}^{|y|}
   \Delta\!\left(
     p_{\theta_0}\!\left(\cdot\mid
       \mathcal C_t^{\mathrm{ref}}(x,g,\tau,y_{<t})\right),
     p_\theta(\cdot\mid x,y_{<t})
   \right)
 \right],
 \label{eq:fixed-opsd}
\end{equation}
where the teacher remains fixed at $\theta_0$ and receives no gradients, even as the student and
its generated responses change throughout training.

\subsection{Self-Refinement and Efficient Reasoning}

Self-training turns a model's own generations into new training data. STaR iteratively trains on
rationales that yield correct answers and uses answer-conditioned rationalization to recover
additional examples, while ReST generates, filters, and reuses model samples as offline data
~\citep{zelikman2022star,gulcehre2023reinforced}. In alignment, SAO generates its own prompts,
responses, and preferences without external annotation~\citep{yin-etal-2026-aligning}, whereas PIKA
uses an external generator and reward model to construct synthetic training
data~\citep{yin2025pika}. Reflexion instead stores verbal feedback to guide
later attempts without updating model parameters~\citep{shinn2023reflexion}, and subsequent work
links verification, backtracking, and subgoal construction to successful self-improvement
~\citep{gandhi2025cognitive}. SRCL follows this filtered self-training perspective but serves a
different purpose: the current checkpoint rewrites its own on-policy response without seeing the
verified solution, and training retains only rewrites that are shorter, naturally terminated,
structurally valid, and answer-correct. SRCL therefore teaches concise successful solutions,
complementing DCE's guidance on how to revise the original response.

Test-time scaling improves accuracy by allocating more computation during inference. Repeated
sampling, search, and compute-aware decoding trade additional inference compute for stronger
performance~\citep{brown2024large,snell2024scaling,wu2025inference}, while input-adaptive methods
allocate that compute according to estimated problem difficulty~\citep{damani2024learning}.
Budget forcing specifically uses continuation cues to prevent early termination and extend a
response to a prescribed budget~\citep{muennighoff2025s1}. SRCL instead changes the model during
post-training so that a single rollout can preserve correct reasoning with fewer tokens. Our
matched-budget controls therefore test whether the gains arise merely from longer generation or
from the learned behavior.

\section{Method}
\label{sec:method}

\begin{figure}[!t]
    \centering
    \includegraphics[width=0.92\textwidth]{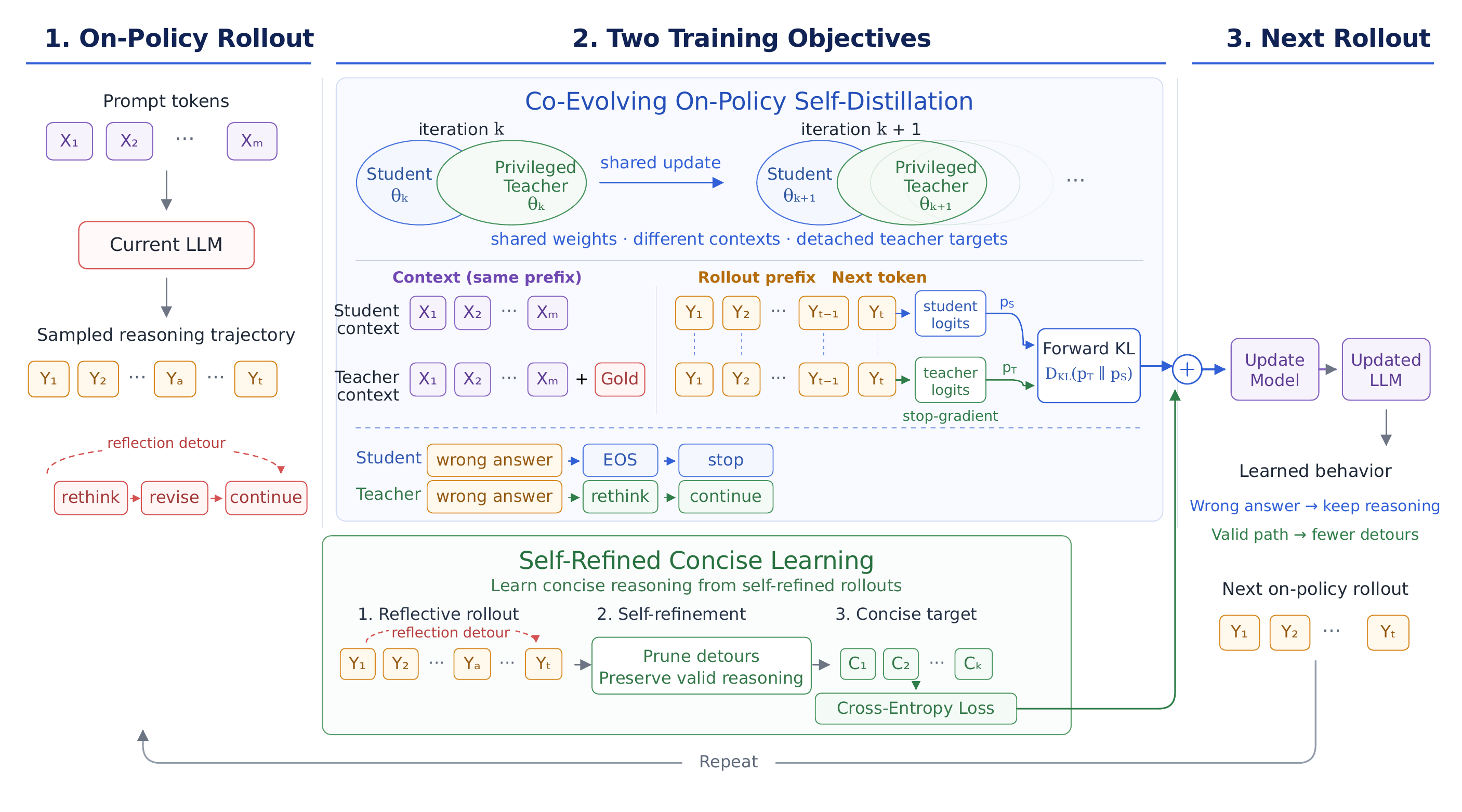}
    \caption{Overview of recursive co-evolution. At round $k$, the current model generates an
    on-policy rollout. DCE supplies gold-conditioned guidance along that rollout, while
    SRCL learns from accepted, shorter self-refinements. Their joint update produces
    $\theta_{k+1}$, which initializes both roles in the next round and carries newly acquired
    revision behavior into future supervision.}
    \label{fig:main-framework}
\end{figure}

To address the limitations of a frozen privileged teacher while controlling the cost of increasingly
long reasoning, our framework combines two complementary training objectives. \textbf{Dynamic
Co-Evolution (DCE)} distills gold-conditioned guidance along the current student's rollout and
refreshes the detached privileged branch from the updated checkpoint each round. \textbf{Self-Refined
Concise Learning (SRCL)} complements this signal by training on shorter, verified refinements of
the same on-policy response. Their joint update produces the checkpoint used as both the student
and privileged teacher in the next round, closing the recursive loop illustrated in
\cref{fig:main-framework}.

\subsection{Dynamic Co-Evolution}

DCE alternates on-policy generation with updates to both the student and its privileged teacher. Let
$\theta_k$ denote the model parameters at training round $k$, with $\theta_0$ representing the
initial checkpoint. Given a problem $x$, the current model samples an on-policy response
\begin{equation}
y^{(k)} \sim p_{\theta_k}(\cdot \mid x).
\end{equation}
From this response, we retain a sequence
$\widetilde y^{(k)}=(\widetilde y^{(k)}_1,\ldots,\widetilde y^{(k)}_m)$ of $m$ tokens on which the
guidance objective is computed. At each token position $t\in\{1,\ldots,m\}$, the student and
privileged teacher score the same response prefix $\widetilde y^{(k)}_{<t}$. The student conditions
on the problem $x$ and this prefix, whereas the privileged teacher additionally receives the
ground-truth solution $g$. Its prompt places the problem $x$ and task instruction in the user turn,
followed in a single assistant turn by $g$, a transition instruction $\tau$ asking the model to
solve the problem using its own approach, and finally $\widetilde y^{(k)}_{<t}$. We denote this privileged context by
$\mathcal C_t^{\mathrm{prev}}(x,g,\tau,\widetilde y^{(k)}_{<t})$,
i.e.\ the concatenation of all privileged inputs up to position $t$.
This ordering is deliberate: placing $g$ in the assistant turn lets the model condition on it as
part of its own reasoning rather than as external user-provided material.
The exact templates and alternative orderings are given in \cref{fig:prompt-serialization}; their
performance is compared in \cref{sec:context-serialization}.

We define the student and privileged-teacher next-token distributions at round $k$ and position $t$ as
\begin{equation}
 \begin{aligned}
 p_{k,t}
   &:=p_{\theta_k}\!\left(
      \cdot\mid x,\widetilde y^{(k)}_{<t}
   \right), \\
 q_{k,t}
   &:=\sg\!\left[
      p_{\theta_k}\!\left(
        \cdot\mid
        \mathcal C_t^{\mathrm{prev}}
        (x,g,\tau,\widetilde y^{(k)}_{<t})
      \right)
   \right].
 \end{aligned}
 \label{eq:dce-distributions}
\end{equation}
Here $p_{k,t}$ is the student distribution, conditioned only on the problem, and $q_{k,t}$ is the
privileged-teacher distribution, which additionally sees the ground-truth solution. Both are
produced by the same current checkpoint $\theta_k$, but gradients are stopped through $q_{k,t}$.
In contrast, standard OPSD differs in two ways: it presents the ground-truth solution as
reference material in the user turn rather than as previously generated reasoning in the
assistant turn, and it keeps the privileged teacher frozen at the initial checkpoint $\theta_0$
throughout training. Let $\Delta(q,p)$ denote a divergence measuring the mismatch between the
teacher and student next-token distributions. DCE minimizes
\begin{equation}
 \mathcal{L}_{G}^{(k)}
 =\mathbb{E}_{(x,g),\,y^{(k)}\sim p_{\theta_k}(\cdot\mid x)}\!\left[
   \frac{1}{m}\sum_{t=1}^{m}
   \Delta\!\left(q_{k,t},p_{k,t}\right)
 \right].
 \label{eq:dynamic-guidance}
\end{equation}
After optimizing this objective at round $k$, the resulting parameters
$\theta_{k+1}$ initialize both the student and privileged teacher for the next round. Consequently,
revision behavior acquired during one round can become part of the gold-conditioned supervision
provided in subsequent rounds, yielding recursive co-evolution. We use Forward KL for $\Delta$ in
our main experiments and compare Reverse KL and JSD in \cref{sec:divergence-control}.

\subsection{Self-Refined Concise Learning}
\label{sec:srcl-method}

DCE improves reasoning by strengthening the model's ability to revisit and revise its own
trajectories. However, stronger revision behavior can also increase inference cost: the model may
perform repeated checks, explore unnecessary branches, or continue reasoning after it has already
reached a correct solution. DCE alone does not explicitly encourage the model to preserve useful
reasoning while removing these redundant steps. We therefore introduce \textbf{Self-Refined
Concise Learning (SRCL)}, which complements DCE by training the model on shorter, verified
refinements of its own on-policy responses.

At training round $k$, let $\mathcal B=\{(x_i,g_i)\}$ denote a minibatch of problems $x_i$ and
their verified solutions $g_i$. For each example $i$, the current checkpoint $\theta_k$ generates
an on-policy response $y_i^{(k)}$. SRCL then asks the same checkpoint to rewrite this response
without access to $g_i$:

\begin{equation}
 c_i^{(k)}
 =R_{\theta_k}\!\left(x_i,y_i^{(k)}\right),
 \label{eq:srcl-rewrite}
\end{equation}
where $R_{\theta_k}$ denotes generation under a refinement prompt that requests a direct,
self-contained solution with unnecessary detours and reflection removed. We retain the refinement
$c_i^{(k)}$ only if it is shorter than the original response $y_i^{(k)}$, terminates naturally,
satisfies structural validity requirements, and produces the correct answer. Let
$A_i^{(k)}\in\{0,1\}$ indicate whether $c_i^{(k)}$ passes all of these criteria; the exact
acceptance rules are given in \cref{app:srcl-acceptance}.

Each accepted refinement becomes a token-level supervised target. SRCL minimizes the
autoregressive cross-entropy over all accepted tokens:
\begin{equation}
 \mathcal{L}_{\mathrm{SRCL}}^{(k)}
 =-\frac{1}{N_k}
 \sum_{i\in\mathcal B} A_i^{(k)}
 \sum_{t=1}^{|c_i^{(k)}|}
 \log p_{\theta_k}\!\left(
   c_{i,t}^{(k)}
   \mid
   x_i,c_{i,<t}^{(k)}
 \right),
 \qquad
 N_k
 =\sum_{i\in\mathcal B}
 A_i^{(k)}|c_i^{(k)}|.
 \label{eq:srcl-loss}
\end{equation}
Here, $i$ indexes examples in the minibatch and $t$ indexes tokens within an accepted refinement.
If no refinement is accepted in a minibatch, such that $N_k=0$, we omit the SRCL loss and update
the model using DCE alone. Importantly, the verified solution $g_i$ is used only to determine
whether a refinement is accepted; it is never provided to the model when generating
$c_i^{(k)}$, preventing ground-truth leakage into the learned rewriting behavior. The exact refinement prompt, structural acceptance criteria, and answer-verifier
implementation are provided in \cref{app:srcl-acceptance,app:prompt-templates}.

\subsection{Joint Optimization}
\label{sec:joint-optimization}

DCE and SRCL provide complementary supervision from the same on-policy response. DCE trains the
model to follow the privileged teacher at prefixes encountered along the current rollout, whereas
SRCL trains on shorter, verified refinements derived from that rollout. We combine the two
objectives as
\begin{equation}
 \mathcal{L}_{\mathrm{DCE+SRCL}}^{(k)}
 =
 \underbrace{\lambda_G
 \mathcal{L}_{G}^{(k)}}_{\text{DCE guidance}}
 +
 \underbrace{\lambda_S
 \mathcal{L}_{\mathrm{SRCL}}^{(k)}}_{\text{SRCL target}}.
 \label{eq:joint-objective}
\end{equation}
Here, $\mathcal{L}_{G}^{(k)}$ is the dynamic guidance loss defined in
\cref{eq:dynamic-guidance}, and $\mathcal{L}_{\mathrm{SRCL}}^{(k)}$ is the self-refinement loss
defined in \cref{eq:srcl-loss}. The coefficients $\lambda_G$ and
$\lambda_S$ control the relative contribution of the two objectives; their empirical
sensitivity is studied in \cref{sec:guidance-weight}. The complete recursive procedure is
summarized in \cref{alg:dce-srcl}.

\begin{algorithm}[t]
\caption{Recursive training with DCE and SRCL}
\label{alg:dce-srcl}
\small
\begin{algorithmic}[1]
\Require Initial checkpoint $\theta_0$; training set
         $\mathcal D=\{(x_i,g_i)\}$; number of rounds $K$;
         weights $\lambda_G,\lambda_S$
\For{$k=0,1,\ldots,K-1$}
    \State Sample a minibatch $\mathcal B\subset\mathcal D$
    \State Initialize $\mathcal B_{\mathrm{SRCL}}\gets\emptyset$
    \ForAll{$(x_i,g_i)\in\mathcal B$}
        \State Sample an on-policy response
        $y_i^{(k)}\sim p_{\theta_k}(\cdot\mid x_i)$
        \State Score the response prefixes with the student and detached
        gold-conditioned privileged teacher
        \State Generate a self-refinement
        $c_i^{(k)}=R_{\theta_k}(x_i,y_i^{(k)})$ without access to $g_i$
        \State Compute the acceptance indicator $A_i^{(k)}$
        \If{$A_i^{(k)}=1$}
            \State Add $(x_i,c_i^{(k)})$ to $\mathcal B_{\mathrm{SRCL}}$
        \EndIf
    \EndFor
    \State Compute $\mathcal L_G^{(k)}$ from the scored on-policy prefixes
    \If{$\mathcal B_{\mathrm{SRCL}}\neq\emptyset$}
        \State Compute $\mathcal L_{\mathrm{SRCL}}^{(k)}$ on
        $\mathcal B_{\mathrm{SRCL}}$
    \Else
        \State Set $\mathcal L_{\mathrm{SRCL}}^{(k)}\gets 0$
    \EndIf
    \State Update
    \[
        \theta_{k+1}\gets
        \operatorname{Update}\!\left(
        \theta_k,\,
        \lambda_G\mathcal L_G^{(k)}
        +\lambda_S\mathcal L_{\mathrm{SRCL}}^{(k)}
        \right)
    \]
    \State Use $\theta_{k+1}$ to initialize both the student and privileged teacher
    for round $k+1$
\EndFor
\Ensure Updated checkpoint $\theta_K$
\end{algorithmic}
\end{algorithm}

\section{Experimental Setup}
\label{sec:experimental-setup}

\subsection{Models, Data, and Metrics}

We adopt Qwen3-1.7B, Qwen3-4B, Qwen3-8B, and Qwen3-14B~\citep{qwen3} as our primary model
family, and Gemma-4-12B-IT in the cross-family study in \cref{sec:gemma12b-transfer}. Both the
problem-only student and gold-conditioned privileged teacher operate in non-thinking mode.
All methods use the same 14,717 problems drawn from the OpenThoughts mathematical-reasoning data
adopted by OPSD~\citep{guha2025openthoughtsdatarecipesreasoning,zhao2026selfdistilled}.
DCE constructs student and privileged views of each problem during training. Complete configurations are reported in
\cref{tab:configuration}.

Evaluation uses all 30 problems from each of AIME 2024, AIME 2025, AIME 2026
~\citep{maaAIME}, and HMMT February 2025~\citep{hmmtArchive}. We sample 12 responses per problem
with a 32K generation cap, temperature 1.0, top-$p=0.8$, top-$k=-1$, and repetition penalty 1.0.

\subsection{Baselines and Controls}

We compare against four training baselines: \textbf{Base}, \textbf{SFT}, \textbf{GRPO}, and
\textbf{OPSD}. \textbf{Base} is the original model evaluated in non-thinking mode.
\textbf{SFT} trains on the problem--solution pairs $(x,g)$ using autoregressive cross-entropy,
while \textbf{GRPO} optimizes an outcome reward based on final-answer correctness.
\textbf{OPSD} is our closest training baseline: it distills next-token predictions from a frozen,
gold-conditioned privileged teacher initialized from the same base checkpoint.

To separate the effect of improved training from that of simply allocating more inference tokens,
we additionally include \textbf{OPSD-TTS} as an inference-time control. Following the
extended-thinking intervention of \citet{ghosal2025thinkingmore}, when an OPSD response terminates
before the target budget, a \texttt{Wait} cue resumes generation until the response reaches an
exact 8K or 16K token budget. OPSD-TTS does not modify the model parameters and therefore tests
whether longer generation alone can account for the gains of DCE+SRCL.

\subsection{Training Protocol}
Across the Qwen3 DCE experiments, we train with AdamW~\citep{loshchilov2017decoupled} in bfloat16 using a
learning rate of $5\times10^{-6}$ and rank-128 LoRA~\citep{hu2022lora}. Each training example
produces a single on-policy response. DCE computes privileged-teacher guidance over the retained
response tokens, while SRCL greedily rewrites the same response and trains only on accepted targets
that are shorter, reflection-free, and answer-correct.

We set the DCE loss weight to
$\lambda_{\mathrm{DCE}}=5\times10^4$ for all Qwen3 models. For SRCL, we use
$\lambda_{\mathrm{SRCL}}=12.5,25,35,$ and $25$ for Qwen3-14B, 8B, 4B, and 1.7B, respectively.
Sensitivity to both coefficients is analyzed in \cref{sec:guidance-weight}. We evaluate checkpoints
at training steps $\{10,20,30,50,100,150,200\}$ when available.

Baseline-specific SFT and GRPO hyperparameters are reported in \cref{tab:configuration}, and
checkpoint-level learning trajectories are provided in
\cref{fig:learning-dynamics,app:full-trajectories}. The Gemma transfer configuration is reported
separately in \cref{app:gemma12b-details}.

\section{Results}
\label{sec:results}

\subsection{Main Experimental Results}

\begin{table}[H]
\centering
\caption{\textbf{Main results.} Cells report Average@12 accuracy (\textbf{Acc.}, \%) and mean
generated tokens (\textbf{Tok.}). Unmarked rows use non-thinking, 32K decoding; gray Base
(thinking)$^{\ddagger}$ rows are cross-mode references excluded from the main ranking.
$\dagger$ marks the inference-only 16K OPSD-TTS control, and blue bold marks the best non-thinking
accuracy within each model size. Complete trajectories appear in
\cref{fig:learning-dynamics,app:full-trajectories}.}
\label{tab:main-results}
\footnotesize
\setlength{\tabcolsep}{7.5pt}
\renewcommand{\arraystretch}{0.98}
\begin{tabular}{@{}l*{5}{rr}@{}}
\toprule
\textbf{Method}
 & \multicolumn{2}{c}{\textbf{AIME24}}
 & \multicolumn{2}{c}{\textbf{AIME25}}
 & \multicolumn{2}{c}{\textbf{AIME26}}
 & \multicolumn{2}{c}{\textbf{HMMT25}}
 & \multicolumn{2}{c}{\textcolor{metablue}{\textbf{Average}}} \\
\cmidrule(lr){2-3}\cmidrule(lr){4-5}\cmidrule(lr){6-7}\cmidrule(lr){8-9}\cmidrule(l){10-11}
 & Acc. & Tok. & Acc. & Tok. & Acc. & Tok. & Acc. & Tok.
 & \textcolor{metablue}{Acc.} & \textcolor{metablue}{Tok.} \\
\midrule
\rowcolor{metablue!7}
\multicolumn{11}{@{}l}{\textcolor{metablue}{\sffamily\bfseries Qwen3--8B}} \\
Base
 & 28.89 & \resulttok{4,706} & 21.39 & \resulttok{3,189}
 & 17.78 & \resulttok{4,631} & 10.00 & \resulttok{3,161} & 19.51 & \resulttok{3,922} \\
SFT
 & 28.33 & \resulttok{4,662} & 21.94 & \resulttok{3,583}
 & 18.06 & \resulttok{5,075} & 11.67 & \resulttok{3,030} & 20.00 & \resulttok{4,088} \\
GRPO
 & 31.11 & \resulttok{4,710} & 22.50 & \resulttok{3,242}
 & 16.39 & \resulttok{4,007} & 11.11 & \resulttok{2,860} & 20.28 & \resulttok{3,705} \\
OPSD
 & 46.11 & \resulttok{5,678} & 28.61 & \resulttok{6,458}
 & 29.72 & \resulttok{6,371} & 16.94 & \resulttok{5,529} & 30.35 & \resulttok{6,009} \\
\rowcolor{metapurple!4}
OPSD-TTS$^{\dagger}$
 & 46.39 & \resulttok{16,384} & 28.89 & \resulttok{16,384}
 & 31.11 & \resulttok{16,384} & 16.67 & \resulttok{16,384} & 30.76 & \resulttok{16,384} \\
\cmidrule(lr){1-11}
\rowcolor{metablue!3}
\textbf{DCE}
 & \bestacc{74.72} & \resulttok{16,869} & 68.61 & \resulttok{18,562}
 & \bestacc{72.78} & \resulttok{17,550} & \bestacc{46.94} & \resulttok{23,203} & 65.76 & \resulttok{19,046} \\
\rowcolor{metateal!4}
\quad + SRCL
 & 73.89 & \resulttok{15,640} & \bestacc{71.94} & \resulttok{17,241}
 & 71.94 & \resulttok{16,158} & 46.11 & \resulttok{21,205} & \bestacc{65.97} & \resulttok{17,561} \\
\specialrule{0.35pt}{0pt}{0pt}
\rowcolor{black!10}
\textit{Base (thinking)}$^{\ddagger}$
 & 75.00 & \resulttok{15,565} & 64.72 & \resulttok{18,784}
 & 64.44 & \resulttok{17,192} & 44.17 & \resulttok{20,940} & 62.08 & \resulttok{18,120} \\
\midrule
\rowcolor{metablue!7}
\multicolumn{11}{@{}l}{\textcolor{metablue}{\sffamily\bfseries Qwen3--4B}} \\
Base
 & 21.39 & \resulttok{4,754} & 20.56 & \resulttok{3,514}
 & 17.22 & \resulttok{4,142} & 11.11 & \resulttok{2,809} & 17.57 & \resulttok{3,805} \\
SFT
 & 23.89 & \resulttok{4,522} & 20.83 & \resulttok{3,388}
 & 19.17 & \resulttok{4,083} & 10.83 & \resulttok{2,880} & 18.68 & \resulttok{3,718} \\
GRPO
 & 25.28 & \resulttok{3,577} & 20.83 & \resulttok{3,082}
 & 16.11 & \resulttok{3,915} & 11.67 & \resulttok{2,876} & 18.47 & \resulttok{3,363} \\
OPSD
 & 30.83 & \resulttok{9,419} & 22.50 & \resulttok{7,970}
 & 24.17 & \resulttok{8,228} & 13.89 & \resulttok{7,782} & 22.85 & \resulttok{8,350} \\
\rowcolor{metapurple!4}
OPSD-TTS$^{\dagger}$
 & 30.28 & \resulttok{16,384} & 22.78 & \resulttok{16,384}
 & 25.56 & \resulttok{16,384} & 13.33 & \resulttok{16,384} & 22.99 & \resulttok{16,384} \\
\cmidrule(lr){1-11}
\rowcolor{metablue!3}
\textbf{DCE}
 & 68.06 & \resulttok{17,777} & 58.89 & \resulttok{19,644}
 & 69.17 & \resulttok{17,944} & 43.89 & \resulttok{22,073} & 60.00 & \resulttok{19,360} \\
\rowcolor{metateal!4}
\quad + SRCL
 & \bestacc{74.17} & \resulttok{15,273} & \bestacc{59.44} & \resulttok{17,466}
 & \bestacc{69.72} & \resulttok{16,181} & \bestacc{44.17} & \resulttok{20,143} & \bestacc{61.88} & \resulttok{17,265} \\
\specialrule{0.35pt}{0pt}{0pt}
\rowcolor{black!10}
\textit{Base (thinking)}$^{\ddagger}$
 & 72.50 & \resulttok{15,013} & 61.39 & \resulttok{18,037}
 & 64.44 & \resulttok{15,996} & 41.67 & \resulttok{18,631} & 60.00 & \resulttok{16,919} \\
\midrule
\rowcolor{metablue!7}
\multicolumn{11}{@{}l}{\textcolor{metablue}{\sffamily\bfseries Qwen3--1.7B}} \\
Base
 & 14.44 & \resulttok{3,464} & 7.78 & \resulttok{2,633}
 & 8.33 & \resulttok{3,892} & 6.11 & \resulttok{2,350} & 9.17 & \resulttok{3,085} \\
SFT
 & 15.00 & \resulttok{3,600} & 8.33 & \resulttok{2,584}
 & 11.11 & \resulttok{3,957} & 6.11 & \resulttok{2,568} & 10.14 & \resulttok{3,177} \\
GRPO
 & 14.44 & \resulttok{3,539} & 11.39 & \resulttok{2,546}
 & 10.00 & \resulttok{3,715} & 7.78 & \resulttok{2,412} & 10.90 & \resulttok{3,053} \\
OPSD
 & 16.11 & \resulttok{6,556} & 7.78 & \resulttok{5,050}
 & 9.72 & \resulttok{6,164} & 7.78 & \resulttok{3,751} & 10.35 & \resulttok{5,380} \\
\rowcolor{metapurple!4}
OPSD-TTS$^{\dagger}$
 & 14.72 & \resulttok{16,384} & 7.50 & \resulttok{16,384}
 & 8.89 & \resulttok{16,384} & 5.83 & \resulttok{16,384} & 9.24 & \resulttok{16,384} \\
\cmidrule(lr){1-11}
\rowcolor{metablue!3}
\textbf{DCE}
 & 31.39 & \resulttok{13,709} & 24.72 & \resulttok{11,848}
 & 20.83 & \resulttok{12,943} & 16.39 & \resulttok{11,898} & 23.33 & \resulttok{12,600} \\
\rowcolor{metateal!4}
\quad + SRCL
 & \bestacc{33.61} & \resulttok{19,849} & \bestacc{28.33} & \resulttok{18,396}
 & \bestacc{28.89} & \resulttok{19,725} & \bestacc{16.67} & \resulttok{20,046} & \bestacc{26.88} & \resulttok{19,504} \\
\specialrule{0.35pt}{0pt}{0pt}
\rowcolor{black!10}
\textit{Base (thinking)}$^{\ddagger}$
 & 48.33 & \resulttok{17,581} & 38.89 & \resulttok{17,559}
 & 36.94 & \resulttok{18,260} & 23.33 & \resulttok{18,533} & 36.88 & \resulttok{17,983} \\
\bottomrule
\end{tabular}
\end{table}

\Cref{tab:main-results} summarizes the main results. DCE substantially improves over the training
baselines across model scales. For Qwen3-8B, DCE+SRCL reaches 65.97\% Average@12, compared with
30.35\% for OPSD and 20.28\% for GRPO. The gains remain substantial with Qwen3-4B, where
DCE+SRCL reaches 61.88\%, compared with 22.85\% for OPSD and 18.47\% for GRPO. DCE alone
achieves similar accuracy with Qwen3-8B (65.76\%) and reaches 60.00\% with Qwen3-4B,
indicating that the primary accuracy gains come from dynamic co-evolution.

SRCL improves the accuracy--length tradeoff at these scales. With Qwen3-8B, adding SRCL reduces
mean output length from 19,046 to 17,561 tokens while maintaining comparable Average@12 accuracy
(65.76\% versus 65.97\%). With Qwen3-4B, SRCL both improves accuracy from 60.00\% to 61.88\%
and reduces mean output length from 19,360 to 17,265 tokens, a 10.82\% reduction. This effect is
not uniform at the smallest scale: with Qwen3-1.7B, DCE+SRCL improves Average@12 from 23.33\%
to 26.88\%, but also increases mean output length from 12,600 to 19,504 tokens.

\begin{figure}[H]
    \centering
    \includegraphics[width=0.82\textwidth]{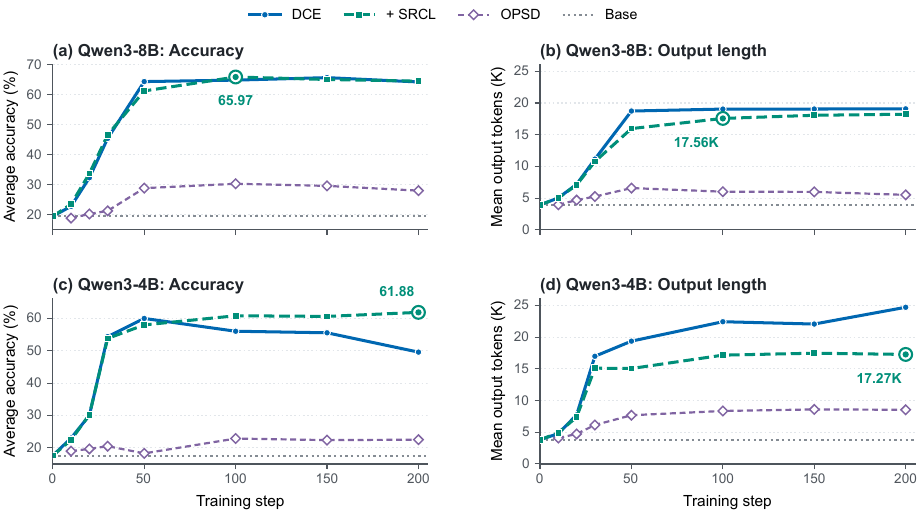}
    \caption{\textbf{Learning dynamics across model scales.} Qwen3-8B (top) and Qwen3-4B
    (bottom) under non-thinking, 32K, Average@12 evaluation. Each row reports Average accuracy
    (left) and mean generated tokens (right). DCE+SRCL peaks at step 100 for 8B and remains stable
    through step 200 for 4B; DCE peaks at steps 150 and 50, respectively.}
    \label{fig:learning-dynamics}
\end{figure}

Longer inference alone does not explain the gains. Forcing OPSD to generate exactly 16K tokens
yields only 30.76\%, 22.99\%, and 9.24\% Average@12 for Qwen3-8B, Qwen3-4B, and Qwen3-1.7B,
respectively. The matched training trajectories in \cref{fig:learning-dynamics} further separate the
effect of SRCL from checkpoint selection: at step 100 with Qwen3-8B, DCE+SRCL reaches 65.97\%
with 17,561 mean tokens, compared with 64.93\% and 19,036 tokens for DCE at the same step. With
Qwen3-4B, DCE peaks earlier and declines after step 50, whereas DCE+SRCL remains near 61\%
through step 200. Complete checkpoint trajectories and budget controls are provided in
\cref{app:full-trajectories,app:qwen17,sec:cueing,tab:opsd-tts-full}.

\FloatBarrier
\subsection{Does Recursive Improvement Transfer Beyond Qwen3?}
\label{sec:gemma12b-transfer}

\begin{table}[H]
\centering
\caption{\textbf{Transfer to Gemma-4-12B-IT.} Cells report Average@12 accuracy (Acc., \%) and
mean generated tokens (Tok.) under non-thinking, 32K decoding. Blue bold marks the highest
accuracy in each column.}
\label{tab:gemma12b-transfer}
\footnotesize
\setlength{\tabcolsep}{1.8pt}
\renewcommand{\arraystretch}{1.04}
\begin{tabular*}{0.99\textwidth}{@{\extracolsep{\fill}}l*{5}{r@{\hspace{1.8pt}}r}@{}}
\toprule
\textbf{Method}
 & \multicolumn{2}{c}{\textbf{AIME24}}
 & \multicolumn{2}{c}{\textbf{AIME25}}
 & \multicolumn{2}{c}{\textbf{AIME26}}
 & \multicolumn{2}{c}{\textbf{HMMT25}}
 & \multicolumn{2}{c}{\textcolor{metablue}{\textbf{Average}}} \\
\cmidrule(lr){2-3}\cmidrule(lr){4-5}\cmidrule(lr){6-7}\cmidrule(lr){8-9}\cmidrule(l){10-11}
 & Acc. & Tok. & Acc. & Tok. & Acc. & Tok. & Acc. & Tok.
 & \textcolor{metablue}{Acc.} & \textcolor{metablue}{Tok.} \\
\midrule
Base
 & 66.67 & \resulttok{6,414} & 51.67 & \resulttok{9,688}
 & 63.33 & \resulttok{7,605} & 43.06 & \resulttok{10,358}
 & 56.18 & \resulttok{8,516} \\
SFT
 & 68.33 & \resulttok{6,676} & 55.83 & \resulttok{8,872}
 & 59.17 & \resulttok{8,476} & 37.22 & \resulttok{11,491}
 & 55.14 & \resulttok{8,878} \\
GRPO
 & 70.83 & \resulttok{5,726} & 51.39 & \resulttok{8,849}
 & 60.56 & \resulttok{8,507} & 39.72 & \resulttok{10,659}
 & 55.63 & \resulttok{8,435} \\
OPSD
 & 64.72 & \resulttok{6,768} & 46.39 & \resulttok{8,865}
 & 53.33 & \resulttok{7,814} & 39.72 & \resulttok{9,001}
 & 51.04 & \resulttok{8,112} \\
\cmidrule(lr){1-11}
\rowcolor{metablue!3}
\textbf{DCE}
 & 71.39 & \resulttok{6,501} & \bestacc{65.00} & \resulttok{8,117}
 & 65.28 & \resulttok{8,724} & 46.39 & \resulttok{11,909}
 & 62.01 & \resulttok{8,812} \\
\rowcolor{metateal!4}
\quad + SRCL
 & \bestacc{72.50} & \resulttok{6,661} & 64.72 & \resulttok{8,341}
 & \bestacc{68.06} & \resulttok{7,965} & \bestacc{49.17} & \resulttok{11,137}
 & \bestacc{63.61} & \resulttok{8,526} \\
\bottomrule
\end{tabular*}
\end{table}

To test whether recursive improvement transfers across model families, we apply our framework to
Gemma-4-12B-IT. As shown in \cref{tab:gemma12b-transfer}, DCE and DCE+SRCL reach 62.01\% and
63.61\% Average@12, respectively, outperforming all non-DCE comparisons. SRCL adds 1.60
percentage points while reducing mean output length from 8,812 to 8,526 tokens. Across the saved
trajectory, DCE+SRCL attains higher accuracy at five of seven checkpoints, ties once, and is both
more accurate and shorter than DCE at four checkpoints. The framework therefore transfers beyond
Qwen3. Detailed configurations and complete training trajectories appear in
\cref{app:gemma12b-details}.

\FloatBarrier
\subsection{What Happens During Recursive Self-Improvement?}
\label{sec:eos-probe}

To characterize recursive self-improvement, we evaluate checkpoints from a Qwen3-8B DCE+SRCL run
on a fixed set of incorrect trajectories. Holding the trajectories
constant isolates two next-token behaviors: whether the model terminates after a wrong solution,
and whether it predicts the reflection cue observed when the solution begins to revise.

For a stored response $y_i^-$ with its terminal EOS removed, the Student scores
$[x_i;y_i^-]$, while the Teacher scores the same response with the verified solution prepended.
At an observed revision point $t$, we instead score the preceding prefix and its actual next cue
$r_{i,t}$, such as \texttt{Wait}. Denote the resulting endpoint and revision contexts by
$c^{\mathrm{end}}$ and $c^{\mathrm{ref}}$. For branch $b\in\{S,T\}$, the two probes are
\begin{equation}
 P_b^{(k)}(\eos)=\frac{1}{|\mathcal E|}\sum_{i\in\mathcal E}
 p_{\theta_k}(\eos\mid c^{\mathrm{end}}_{i,b}),\qquad
 P_b^{(k)}(r)=\frac{1}{|\mathcal W|}\sum_{(i,t)\in\mathcal W}
 p_{\theta_k}(r_{i,t}\mid c^{\mathrm{ref}}_{i,t,b}).
 \label{eq:reflection-probe}
\end{equation}
Both values come directly from the full next-token distribution. The reflection probe scores the
cue actually present in the stored trajectory rather than summing over a hand-built lexicon.

\begin{figure}[H]
    \centering
    \includegraphics[width=0.98\textwidth]{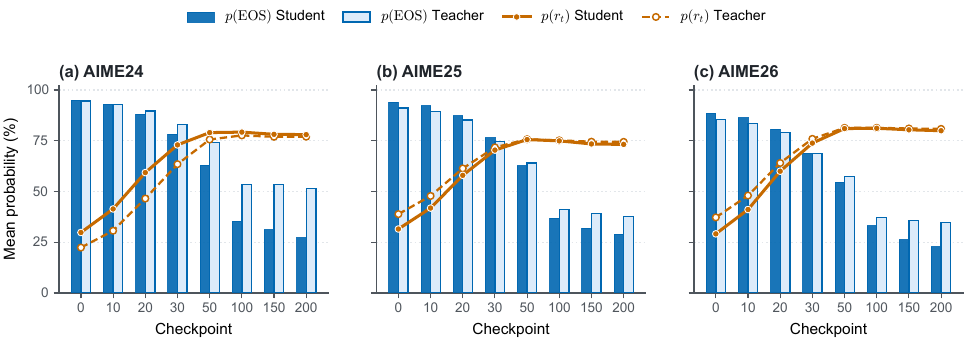}
    \caption{Termination and reflection during recursive self-improvement. On fixed incorrect
    trajectories, endpoint $p(\eos)$ decreases while the probability of the observed reflection
    token $p(r_t)$ increases. Blue bars show EOS; orange curves show $r_t$; dark/solid and
    light/dashed marks denote the Student and gold-conditioned Teacher.}
    \label{fig:eos-probe-curves}
\end{figure}

\Cref{fig:eos-probe-curves} shows the same qualitative transition on all three benchmarks.
Macro-averaged across the cohorts, Student/Teacher endpoint $p(\eos)$ falls from
92.6\%/90.4\% at initialization to 26.3\%/41.3\% at step 200. Over the same interval,
$p(r_t)$ rises from 30.1\%/32.8\% to 77.0\%/77.4\%, with most of the increase occurring by
steps 50--100. Recursive training therefore changes not only the student policy but also the
privileged branch used to supervise the next round: both become less likely to stop after a wrong
solution and more likely to support the observed reflection cue. This shared shift suggests that
co-evolution transfers emerging revision behavior into the privileged teacher, complementing the
frozen-teacher evidence in \cref{fig:frozen-teacher-probe}; complete AIME26 probe values are reported in
\cref{tab:eos-probe-values}.

\FloatBarrier
\subsection{How Do Different Loss Weights Affect Performance?}
\label{sec:guidance-weight}

To explore the distinct roles of the two training objectives, we vary their coefficients
separately on Qwen3-8B. The DCE weight $\lambda_G$ controls the strength of privileged next-token
guidance, whereas the SRCL weight $\lambda_S$ controls the contribution of accepted concise
rewrites. In each sweep, the other coefficient and all remaining training and evaluation settings
are fixed. \Cref{tab:loss-weight-sensitivity} reports the resulting task accuracy and output length
for each setting.

\begin{table}[H]
\centering
\caption{Qwen3-8B sensitivity to the DCE guidance weight $\lambda_G$ and SRCL weight $\lambda_S$.
The other coefficient remains fixed in each sweep. Results report accuracy (Acc., \%) and mean
generated tokens (Tok.) under non-thinking, 32K, Average@12 evaluation. Blue bold marks the best
accuracy within each block; pale teal marks the main setting.}
\label{tab:loss-weight-sensitivity}
\footnotesize
\setlength{\tabcolsep}{1.65pt}
\renewcommand{\arraystretch}{0.98}
\begin{tabular*}{0.99\textwidth}{@{\extracolsep{\fill}}l*{5}{r@{\hspace{1.8pt}}r}@{}}
\toprule
\textbf{Weight}
 & \multicolumn{2}{c}{\textbf{AIME24}} & \multicolumn{2}{c}{\textbf{AIME25}}
 & \multicolumn{2}{c}{\textbf{AIME26}} & \multicolumn{2}{c}{\textbf{HMMT25}}
 & \multicolumn{2}{c}{\textcolor{metablue}{\textbf{Average}}} \\
\cmidrule(lr){2-3}\cmidrule(lr){4-5}\cmidrule(lr){6-7}\cmidrule(lr){8-9}\cmidrule(l){10-11}
 & Acc. & Tok. & Acc. & Tok. & Acc. & Tok. & Acc. & Tok.
 & \textcolor{metablue}{Acc.} & \textcolor{metablue}{Tok.} \\
\midrule
\rowcolor{metablue!7}
\multicolumn{11}{@{}l}{\textcolor{metablue}{\sffamily\bfseries DCE guidance weight $\lambda_G$}} \\
$1\times10^4$
 & \bestacc{76.67} & \resulttok{16,475} & 66.94 & \resulttok{18,539}
 & 71.39 & \resulttok{17,108} & \bestacc{47.22} & \resulttok{22,708}
 & 65.56 & \resulttok{18,708} \\
$2.5\times10^4$
 & 74.17 & \resulttok{16,310} & 69.72 & \resulttok{17,739}
 & 70.83 & \resulttok{17,006} & 45.56 & \resulttok{21,784}
 & 65.07 & \resulttok{18,210} \\
\rowcolor{metateal!5}
\textbf{$5\times10^4$}
 & 73.89 & \resulttok{15,640} & \bestacc{71.94} & \resulttok{17,241}
 & \bestacc{71.94} & \resulttok{16,158} & 46.11 & \resulttok{21,205}
 & \bestacc{65.97} & \resulttok{17,561} \\
$1\times10^5$
 & 73.89 & \resulttok{17,220} & 66.94 & \resulttok{19,219}
 & 71.11 & \resulttok{17,968} & 43.89 & \resulttok{22,837}
 & 63.96 & \resulttok{19,311} \\
\midrule
\rowcolor{metablue!7}
\multicolumn{11}{@{}l}{\textcolor{metablue}{\sffamily\bfseries SRCL weight $\lambda_S$}} \\
12.5
 & 74.17 & \resulttok{18,496} & 67.50 & \resulttok{20,167}
 & 72.50 & \resulttok{19,109} & 45.83 & \resulttok{24,069}
 & 65.00 & \resulttok{20,460} \\
18.75
 & 71.67 & \resulttok{17,805} & 66.39 & \resulttok{19,172}
 & \bestacc{73.33} & \resulttok{18,034} & 44.44 & \resulttok{23,020}
 & 63.96 & \resulttok{19,508} \\
\rowcolor{metateal!5}
\textbf{25}
 & 73.89 & \resulttok{15,640} & \bestacc{71.94} & \resulttok{17,241}
 & 71.94 & \resulttok{16,158} & 46.11 & \resulttok{21,205}
 & \bestacc{65.97} & \resulttok{17,561} \\
30
 & 72.78 & \resulttok{18,004} & 68.61 & \resulttok{19,436}
 & 72.50 & \resulttok{18,544} & \bestacc{47.22} & \resulttok{23,358}
 & 65.28 & \resulttok{19,836} \\
32.5
 & \bestacc{75.56} & \resulttok{16,524} & 63.89 & \resulttok{18,977}
 & 68.33 & \resulttok{17,914} & 45.83 & \resulttok{22,424}
 & 63.40 & \resulttok{18,960} \\
50
 & 74.44 & \resulttok{16,874} & 68.33 & \resulttok{19,020}
 & 70.00 & \resulttok{18,369} & 46.39 & \resulttok{22,932}
 & 64.79 & \resulttok{19,299} \\
500
 & 74.17 & \resulttok{15,039} & 63.61 & \resulttok{17,293}
 & 71.94 & \resulttok{15,412} & 43.89 & \resulttok{20,952}
 & 63.40 & \resulttok{17,174} \\
\bottomrule
\end{tabular*}
\end{table}

Both sweeps identify a broad but non-monotonic operating region. Across the tested $\lambda_G$
range, Average accuracy varies only from 63.96\% to 65.97\%; $\lambda_G=5\times10^4$ gives
both the highest Average and the lowest mean length, although $1\times10^4$ is stronger on AIME24
and HMMT25. For SRCL, $\lambda_S=25$ likewise gives the highest Average, 65.97\%, while the other
settings remain within 2.57 points despite spanning a forty-fold range. These results indicate
local robustness, not a monotonic or scale-independent optimum; the corresponding 4B sweep is
reported in \cref{tab:srcl-coefficient-4b}. They also clarify that the objectives are
complementary: when DCE is removed, SRCL alone collapses to 2.36\%/1,116 tokens at 8B and
0.76\%/531 tokens at 4B. Concise self-refinement therefore improves the frontier only when paired
with dynamic guidance that develops the underlying revision capability.

\subsection{How Does the Guidance Objective Influence Performance?}
\label{sec:divergence-control}

To examine how the direction of distillation affects recursive improvement, we compare the best
observed Qwen3-8B DCE+SRCL runs using Forward KL $D_{\mathrm{KL}}(q\|p)$, Reverse KL
$D_{\mathrm{KL}}(p\|q)$, and JSD
$\tfrac12D_{\mathrm{KL}}(q\|m)+\tfrac12D_{\mathrm{KL}}(p\|m)$, where $m=(q+p)/2$, under the same
evaluation protocol.

\begin{table}[!t]
\centering
\caption{Effect of the distillation objective on Qwen3-8B DCE+SRCL under non-thinking, 32K,
Average@12 evaluation. Accuracy (Acc., \%) and mean generated tokens (Tok.) are reported
separately.}
\label{tab:divergence-control}
\footnotesize
\setlength{\tabcolsep}{1.55pt}
\renewcommand{\arraystretch}{1.12}
\begin{tabular*}{0.99\textwidth}{@{\extracolsep{\fill}}l*{5}{r@{\hspace{1.5pt}}r}@{}}
\toprule
\textbf{Objective}
 & \multicolumn{2}{c}{\textbf{AIME24}} & \multicolumn{2}{c}{\textbf{AIME25}}
 & \multicolumn{2}{c}{\textbf{AIME26}} & \multicolumn{2}{c}{\textbf{HMMT25}}
 & \multicolumn{2}{c}{\textcolor{metablue}{\textbf{Average}}} \\
\cmidrule(lr){2-3}\cmidrule(lr){4-5}\cmidrule(lr){6-7}\cmidrule(lr){8-9}\cmidrule(l){10-11}
 & Acc. & Tok. & Acc. & Tok. & Acc. & Tok. & Acc. & Tok.
 & \textcolor{metablue}{Acc.} & \textcolor{metablue}{Tok.} \\
\midrule
\rowcolor{metateal!4}
\textbf{Forward KL}
 & \bestacc{73.89} & \resulttok{15,640} & \bestacc{71.94} & \resulttok{17,241}
 & \bestacc{71.94} & \resulttok{16,158} & \bestacc{46.11} & \resulttok{21,205}
 & \bestacc{65.97} & \resulttok{17,561} \\
\rowcolor{metapurple!5}
Reverse KL
 & 69.44 & \resulttok{21,818} & 63.61 & \resulttok{23,800}
 & 67.50 & \resulttok{22,574} & 41.11 & \resulttok{26,801}
 & 60.42 & \resulttok{23,748} \\
JSD
 & 25.83 & \resulttok{4,776} & 18.33 & \resulttok{3,014}
 & 16.39 & \resulttok{4,453} & 11.39 & \resulttok{2,823}
 & 17.99 & \resulttok{3,767} \\
\bottomrule
\end{tabular*}
\end{table}

The results reflect the asymmetry of the three objectives. Forward KL weights discrepancies by the
teacher distribution, preserving a strong signal for corrections that the student underweights.

Reverse KL instead weights the mismatch by
the student distribution; corrections that the student rarely considers contribute less, favoring
its existing modes over missing teacher-supported alternatives. JSD is symmetric and bounded,
which can further weaken directional transfer when the two distributions differ substantially.
Consistent with this interpretation, Forward KL reaches 65.97\% with 17,561 tokens, whereas Reverse
KL is 5.55 percentage points lower while using 6,187 more tokens. JSD reaches only 17.99\% with
3,767 tokens, suggesting premature shortening rather than useful concision. Forward KL is therefore
the most effective of the tested objectives for transferring privileged revision guidance.

\newcommand{\budgettable}{%
\begin{table}[!t]
\centering
\caption{Qwen3-8B performance under matched 8K and 16K budgets. Truncate cuts a 32K response at
the budget; Cue inserts a continuation instruction before the same cap; TTS extends OPSD to the
exact budget. Results report accuracy (Acc., \%) and mean output tokens (Tok.). Blue bold marks the
stronger result within each matched comparison.}
\label{tab:budget-full}
\scriptsize
\setlength{\tabcolsep}{1.15pt}
\renewcommand{\arraystretch}{0.94}
\begin{tabular*}{\textwidth}{@{\extracolsep{\fill}}ll*{5}{r@{\hspace{1.1pt}}r}@{}}
\toprule
\textbf{Method} & \textbf{Condition}
 & \multicolumn{2}{c}{\textbf{AIME24}} & \multicolumn{2}{c}{\textbf{AIME25}}
 & \multicolumn{2}{c}{\textbf{AIME26}} & \multicolumn{2}{c}{\textbf{HMMT25}}
 & \multicolumn{2}{c}{\textcolor{metablue}{\textbf{Average}}} \\
\cmidrule(lr){3-4}\cmidrule(lr){5-6}\cmidrule(lr){7-8}\cmidrule(lr){9-10}\cmidrule(l){11-12}
 & & Acc. & Tok. & Acc. & Tok. & Acc. & Tok. & Acc. & Tok.
 & \textcolor{metablue}{Acc.} & \textcolor{metablue}{Tok.} \\
\midrule
\multicolumn{12}{@{}l}{\textcolor{metablue}{\sffamily\bfseries 8K target}} \\
OPSD & Standard
 & 45.00 & \resulttok{4,148} & 27.50 & \resulttok{4,217}
 & 28.89 & \resulttok{4,500} & 15.28 & \resulttok{4,144}
 & \bestacc{29.17} & \resulttok{4,252} \\
\rowcolor{metapurple!5}
OPSD & \textcolor{metapurple}{\bfseries TTS}
 & 43.06 & \resulttok{8,192} & 27.50 & \resulttok{8,192}
 & 28.33 & \resulttok{8,192} & 15.00 & \resulttok{8,192}
 & 28.47 & \resulttok{8,192} \\
\cmidrule(lr){1-12}
DCE & Truncate
 & 35.56 & \resulttok{7,740} & 23.33 & \resulttok{7,736}
 & 28.06 & \resulttok{7,751} & 10.56 & \resulttok{7,952}
 & 24.38 & \resulttok{7,795} \\
\rowcolor{metateal!7}
DCE & \textcolor{metateal}{\bfseries Cue}
 & 44.72 & \resulttok{7,688} & 30.56 & \resulttok{7,695}
 & 36.39 & \resulttok{7,680} & 12.50 & \resulttok{7,921}
 & \bestacc{31.04} & \resulttok{7,746} \\
DCE+SRCL & Truncate
 & 42.50 & \resulttok{7,400} & 29.44 & \resulttok{7,493}
 & 31.39 & \resulttok{7,574} & 10.83 & \resulttok{7,792}
 & 28.54 & \resulttok{7,565} \\
\rowcolor{metateal!7}
DCE+SRCL & \textcolor{metateal}{\bfseries Cue}
 & 47.22 & \resulttok{7,339} & 35.83 & \resulttok{7,444}
 & 40.83 & \resulttok{7,492} & 16.39 & \resulttok{7,727}
 & \bestacc{35.07} & \resulttok{7,500} \\
\midrule
\multicolumn{12}{@{}l}{\textcolor{metablue}{\sffamily\bfseries 16K target}} \\
OPSD & Standard
 & 46.11 & \resulttok{4,965} & 28.61 & \resulttok{5,236}
 & 29.72 & \resulttok{5,374} & 16.67 & \resulttok{4,862}
 & 30.28 & \resulttok{5,109} \\
\rowcolor{metapurple!5}
OPSD & \textcolor{metapurple}{\bfseries TTS}
 & 46.39 & \resulttok{16,384} & 28.89 & \resulttok{16,384}
 & 31.11 & \resulttok{16,384} & 16.67 & \resulttok{16,384}
 & \bestacc{30.76} & \resulttok{16,384} \\
\cmidrule(lr){1-12}
DCE & Truncate
 & 57.78 & \resulttok{12,535} & 51.67 & \resulttok{13,298}
 & 56.67 & \resulttok{12,824} & 28.61 & \resulttok{14,911}
 & 48.68 & \resulttok{13,392} \\
\rowcolor{metateal!7}
DCE & \textcolor{metateal}{\bfseries Cue}
 & 67.22 & \resulttok{12,258} & 54.17 & \resulttok{13,110}
 & 64.17 & \resulttok{12,804} & 37.50 & \resulttok{14,512}
 & \bestacc{55.76} & \resulttok{13,171} \\
DCE+SRCL & Truncate
 & 61.67 & \resulttok{11,576} & 54.44 & \resulttok{12,605}
 & 58.89 & \resulttok{12,302} & 33.06 & \resulttok{14,411}
 & 52.01 & \resulttok{12,724} \\
\rowcolor{metateal!7}
DCE+SRCL & \textcolor{metateal}{\bfseries Cue}
 & 70.83 & \resulttok{11,424} & 58.89 & \resulttok{12,350}
 & 60.83 & \resulttok{12,036} & 40.00 & \resulttok{13,817}
 & \bestacc{57.64} & \resulttok{12,407} \\
\bottomrule
\end{tabular*}
\end{table}
}

\newcommand{\inferenceanalyses}{%
\subsection{How Does the Method Perform under Tight Reasoning Budgets?}
\label{sec:cueing}

A practical question is whether the learned revision behavior remains useful under tight reasoning
budgets. We therefore evaluate Qwen3-8B with total output budgets of 8K and 16K
using three inference-only controls. \emph{Truncate} directly cuts the response generated under the
original 32K setting at the target budget. \emph{Cue} first generates 7K tokens for an 8K budget
(14K for 16K), appends a short instruction to continue reasoning, and generates the remainder
within the same total cap. \emph{TTS} adapts the
test-time-scaling continuation procedure of \citet{ghosal2025thinkingmore}: when an OPSD response
ends early, it appends a \texttt{Wait} cue and continues until the output reaches the exact target
length. These controls modify decoding only; model parameters remain fixed.

\budgettable

\Cref{tab:budget-full} shows that simply forcing more tokens is ineffective: at 8K, TTS lowers
OPSD from 29.17\% to 28.47\% while nearly doubling output length, and at 16K it gains only 0.48
points while adding 11,275 tokens. Cue, by contrast, improves every matched DCE setting while
slightly reducing mean output. At 8K, DCE rises from 24.38\% to 31.04\% and DCE+SRCL from
28.54\% to 35.07\%; at 16K, the corresponding gains are 48.68\% to 55.76\% and 52.01\% to
57.64\%. Thus the learned revision behavior remains useful under tight budgets, whereas length
alone does not explain the gain. Complete TTS results across model sizes appear in
\cref{tab:opsd-tts-full}.

\subsection{How Sensitive Is Performance to Decoding Hyperparameters?}
\label{sec:decoding-sensitivity}

\begin{table}[!t]
\centering
\caption{Qwen3-8B decoding sensitivity (accuracy / mean tokens). Temperature varies at
$\rho=1.0$ and repetition penalty at $T=1.0$. Pale blue marks the default; blue bold marks the
row best.}
\label{tab:decoding-sensitivity}
\footnotesize
\renewcommand{\arraystretch}{1.10}
\begin{minipage}{0.98\textwidth}
\makebox[\linewidth][l]{\sffamily\bfseries\textcolor{metablue}{(a) Temperature sweep}
\;($\rho=1.0$)}
\par\vspace{3pt}
\centering
\begin{tabular*}{\linewidth}{@{\extracolsep{\fill}}l
 r@{\hspace{2.2pt}}r
 r@{\hspace{2.2pt}}r
 >{\columncolor{metablue!5}}r@{\hspace{2.2pt}}>{\columncolor{metablue!5}}r
 r@{\hspace{2.2pt}}r@{}}
\toprule
\textbf{Benchmark}
 & \multicolumn{2}{c}{$T=0.6$} & \multicolumn{2}{c}{$T=0.8$}
 & \multicolumn{2}{>{\columncolor{metablue!5}}c}{\textcolor{metablue}{\bfseries $T=1.0$}}
 & \multicolumn{2}{c}{$T=1.2$} \\
\cmidrule(lr){2-3}\cmidrule(lr){4-5}\cmidrule(lr){6-7}\cmidrule(l){8-9}
 & Acc. & Tok. & Acc. & Tok. & Acc. & Tok. & Acc. & Tok. \\
\midrule
AIME24 & \bestacc{75.28} & \resulttok{15,155} & 74.17 & \resulttok{15,485}
 & 73.89 & \resulttok{15,640} & 74.72 & \resulttok{15,325} \\
AIME25 & 69.44 & \resulttok{17,530} & 67.78 & \resulttok{17,990}
 & \bestacc{71.94} & \resulttok{17,241} & 71.67 & \resulttok{17,342} \\
AIME26 & 68.33 & \resulttok{16,883} & 68.61 & \resulttok{16,584}
 & \bestacc{71.94} & \resulttok{16,158} & 70.28 & \resulttok{16,462} \\
HMMT25 & 44.72 & \resulttok{21,723} & \bestacc{46.67} & \resulttok{21,522}
 & 46.11 & \resulttok{21,205} & 45.00 & \resulttok{21,395} \\
\midrule
\textcolor{metablue}{\bfseries Average} & 64.44 & \resulttok{17,823} & 64.31 & \resulttok{17,895}
 & \bestacc{65.97} & \resulttok{17,561} & 65.42 & \resulttok{17,631} \\
\bottomrule
\end{tabular*}
\par\vspace{8pt}
\makebox[\linewidth][l]{\sffamily\bfseries\textcolor{metateal}{(b) Repetition-penalty sweep}
\;($T=1.0$)}
\par\vspace{3pt}
\centering
\begin{tabular*}{\linewidth}{@{\extracolsep{\fill}}l
 >{\columncolor{metablue!5}}r@{\hspace{2.2pt}}>{\columncolor{metablue!5}}r
 r@{\hspace{2.2pt}}r
 r@{\hspace{2.2pt}}r
 r@{\hspace{2.2pt}}r@{}}
\toprule
\textbf{Benchmark}
 & \multicolumn{2}{>{\columncolor{metablue!5}}c}{\textcolor{metablue}{\bfseries $\rho=1.00$}}
 & \multicolumn{2}{c}{$\rho=1.02$} & \multicolumn{2}{c}{$\rho=1.08$}
 & \multicolumn{2}{c}{$\rho=1.10$} \\
\cmidrule(lr){2-3}\cmidrule(lr){4-5}\cmidrule(lr){6-7}\cmidrule(l){8-9}
 & Acc. & Tok. & Acc. & Tok. & Acc. & Tok. & Acc. & Tok. \\
\midrule
AIME24 & 73.89 & \resulttok{15,640} & \bestacc{78.06} & \resulttok{15,386}
 & 77.50 & \resulttok{15,317} & 75.56 & \resulttok{14,876} \\
AIME25 & \bestacc{71.94} & \resulttok{17,241} & 68.61 & \resulttok{17,701}
 & 70.00 & \resulttok{16,909} & 71.39 & \resulttok{16,779} \\
AIME26 & 71.94 & \resulttok{16,158} & 72.78 & \resulttok{16,591}
 & \bestacc{73.33} & \resulttok{15,813} & 71.39 & \resulttok{16,139} \\
HMMT25 & 46.11 & \resulttok{21,205} & 47.50 & \resulttok{21,390}
 & \bestacc{47.78} & \resulttok{20,652} & 47.22 & \resulttok{20,962} \\
\midrule
\textcolor{metablue}{\bfseries Average} & 65.97 & \resulttok{17,561} & 66.74 & \resulttok{17,767}
 & \bestacc{67.15} & \resulttok{17,173} & 66.39 & \resulttok{17,189} \\
\bottomrule
\end{tabular*}
\end{minipage}
\end{table}

Finally, we test sensitivity to two common decoding choices under the non-thinking, 32K,
Average@12 protocol. \Cref{tab:decoding-sensitivity} shows that the default $T=1.0$ gives the best
Average among the tested temperatures (65.97\%/17,561 tokens), while $\rho=1.08$ is locally best
among the tested repetition penalties (67.15\%/17,173), gaining 1.18 points with 388 fewer tokens.
We retain default decoding for the main comparisons and report the repetition-penalty result as a
decoding-sensitivity analysis.
}

\FloatBarrier
\subsection{Does the Privileged Teacher Need to Co-Evolve?}
\label{sec:teacher-ablation}

\begin{figure}[H]
    \centering
    \includegraphics[width=0.76\textwidth]{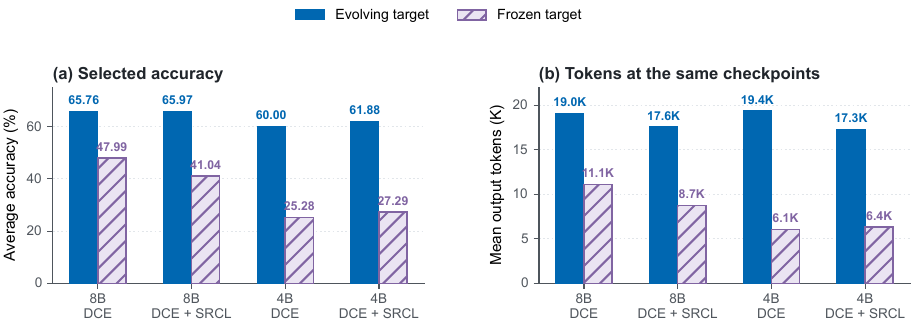}
    \caption{Teacher co-evolution at 8B and 4B. Bars compare evolving and frozen targets for DCE
    and DCE+SRCL in four-task Average accuracy (left) and mean output length (right).}
    \label{fig:teacher-bars}
\end{figure}

To isolate the role of teacher refresh, we compare co-evolving and frozen gold-conditioned
teachers under the same on-policy pipeline, both with and without SRCL. \Cref{fig:teacher-bars}
summarizes the accuracy--length comparison; complete trajectories and task-level results appear in
\cref{app:teacher-trajectories,fig:teacher-ablation,tab:teacher-best}.

Co-evolution improves both configurations. At 8B and 4B, DCE reaches 65.76\% and 60.00\%, versus
47.99\% and 25.28\% when frozen; DCE+SRCL reaches 65.97\% and 61.88\%, versus 41.04\% and
27.29\%. The shorter frozen runs incur large accuracy losses and regress after brief initial gains,
indicating premature termination. Continual refresh instead sustains emerging revision behavior and
keeps gold-conditioned guidance aligned with the evolving model.

\FloatBarrier
\subsection{Which Representation of the Verified Solution Works Better?}
\label{sec:context-serialization}

Beyond teacher refresh, privileged-solution placement also affects signal quality. Assistant-side
prefill places the solution in preceding model-generated context, whereas user-side instruction-last
treats it as external material. Holding all else fixed, assistant-side prefill improves Average by
9.79, 6.39, and 14.66 points at 8B, 4B, and 1.7B and is 780 tokens shorter at 8B
(\cref{tab:context-serialization}). A matched 8B probe localizes its strongest effect near the start
of $Y_0$, whereas user-side effects persist later (\cref{tab:context-signal}), consistent with route
initialization. Placement alone is insufficient: reference-last performs better under a frozen
teacher (\cref{tab:context-teacher-placement}). The best result combines assistant-side conditioning
with continual refresh; \cref{fig:prompt-serialization} gives the exact prompt orderings.

\enlargethispage{1.5\baselineskip}
\begin{table}[H]
\centering
\caption{Privileged-solution placement across model scales under non-thinking, 32K, Average@12
evaluation. Only prompt serialization changes within each model; cells report accuracy (Acc., \%)
and mean generated tokens (Tok.).}
\label{tab:context-serialization}
\footnotesize
\setlength{\tabcolsep}{1.8pt}
\renewcommand{\arraystretch}{0.88}
\begin{tabular*}{0.99\textwidth}{@{\extracolsep{\fill}}l*{5}{r@{\hspace{1.8pt}}r}@{}}
\toprule
\textbf{Representation}
 & \multicolumn{2}{c}{\textbf{AIME24}} & \multicolumn{2}{c}{\textbf{AIME25}}
 & \multicolumn{2}{c}{\textbf{AIME26}} & \multicolumn{2}{c}{\textbf{HMMT25}}
 & \multicolumn{2}{c}{\textcolor{metablue}{\textbf{Average}}} \\
\cmidrule(lr){2-3}\cmidrule(lr){4-5}\cmidrule(lr){6-7}\cmidrule(lr){8-9}\cmidrule(l){10-11}
 & Acc. & Tok. & Acc. & Tok. & Acc. & Tok. & Acc. & Tok.
 & \textcolor{metablue}{Acc.} & \textcolor{metablue}{Tok.} \\
\midrule
\rowcolor{metablue!7}
\multicolumn{11}{@{}l}{\textcolor{metablue}{\sffamily\bfseries Qwen3--8B}} \\
\rowcolor{metateal!6}
\textbf{Assistant-side prefill}
 & \bestacc{73.89} & \resulttok{15,640} & \bestacc{71.94} & \resulttok{17,241}
 & \bestacc{71.94} & \resulttok{16,158} & \bestacc{46.11} & \resulttok{21,205}
 & \bestacc{65.97} & \resulttok{17,561} \\
User-side instruction-last
 & 66.11 & \resulttok{16,982} & 61.94 & \resulttok{18,423}
 & 62.50 & \resulttok{16,336} & 34.17 & \resulttok{21,625}
 & 56.18 & \resulttok{18,341} \\
\midrule
\rowcolor{metablue!7}
\multicolumn{11}{@{}l}{\textcolor{metablue}{\sffamily\bfseries Qwen3--4B}} \\
\rowcolor{metateal!6}
\textbf{Assistant-side prefill}
 & \bestacc{74.17} & \resulttok{15,273} & \bestacc{59.44} & \resulttok{17,466}
 & \bestacc{69.72} & \resulttok{16,181} & \bestacc{44.17} & \resulttok{20,143}
 & \bestacc{61.88} & \resulttok{17,265} \\
User-side instruction-last
 & 65.83 & \resulttok{13,894} & 57.22 & \resulttok{15,498}
 & 59.17 & \resulttok{13,156} & 39.72 & \resulttok{17,741}
 & 55.49 & \resulttok{15,072} \\
\midrule
\rowcolor{metablue!7}
\multicolumn{11}{@{}l}{\textcolor{metablue}{\sffamily\bfseries Qwen3--1.7B}} \\
\rowcolor{metateal!6}
\textbf{Assistant-side prefill}
 & \bestacc{33.61} & \resulttok{19,849} & \bestacc{28.33} & \resulttok{18,396}
 & \bestacc{28.89} & \resulttok{19,725} & \bestacc{16.67} & \resulttok{20,046}
 & \bestacc{26.88} & \resulttok{19,504} \\
User-side instruction-last
 & 15.00 & \resulttok{7,274} & 13.33 & \resulttok{5,594}
 & 11.94 & \resulttok{7,797} & 8.61 & \resulttok{5,040}
 & 12.22 & \resulttok{6,426} \\
\bottomrule
\end{tabular*}
\end{table}

\newcommand{\srclweightappendix}{%
\section{Complete SRCL-Weight Results}
\label{app:srcl-weight}

The main text reports the four-task Average for each coefficient. For completeness,
\cref{tab:srcl-coefficient} provides the corresponding task-level results.
\begin{table}[H]
\centering
\caption{SRCL-coefficient sensitivity under non-thinking, 32K, Average@12 evaluation. Each
row reports one coefficient setting. Accuracy (Acc., \%) and mean generated
tokens (Tok.) are adjacent. Within each model size and metric, blue bold marks the highest
accuracy; pale teal and bold $\lambda_S$/Step identify the coefficients used for the main reported
checkpoints.}
\label{tab:srcl-coefficient}
\footnotesize
\setlength{\tabcolsep}{1.8pt}
\renewcommand{\arraystretch}{1.08}
\begin{tabular*}{0.99\textwidth}{@{\extracolsep{\fill}}lr*{5}{r@{\hspace{2.2pt}}r}@{}}
\toprule
\textbf{$\lambda_S$} & \textbf{Step}
 & \multicolumn{2}{c}{\textbf{AIME24}} & \multicolumn{2}{c}{\textbf{AIME25}}
 & \multicolumn{2}{c}{\textbf{AIME26}} & \multicolumn{2}{c}{\textbf{HMMT25}}
 & \multicolumn{2}{c}{\textcolor{metablue}{\textbf{Average}}} \\
\cmidrule(lr){3-4}\cmidrule(lr){5-6}\cmidrule(lr){7-8}\cmidrule(lr){9-10}\cmidrule(l){11-12}
 & & Acc. & Tok. & Acc. & Tok. & Acc. & Tok. & Acc. & Tok.
 & \textcolor{metablue}{Acc.} & \textcolor{metablue}{Tok.} \\
\midrule
\rowcolor{metablue!7}
\multicolumn{12}{@{}l}{\textcolor{metablue}{\sffamily\bfseries Qwen3--8B}} \\
12.5 & 150 & 74.17 & \resulttok{18,496} & 67.50 & \resulttok{20,167}
 & 72.50 & \resulttok{19,109} & 45.83 & \resulttok{24,069} & 65.00 & \resulttok{20,460} \\
18.75 & 100 & 71.67 & \resulttok{17,805} & 66.39 & \resulttok{19,172}
 & \bestacc{73.33} & \resulttok{18,034} & 44.44 & \resulttok{23,020} & 63.96 & \resulttok{19,508} \\
\rowcolor{metateal!5}
\textbf{25} & \textbf{100} & 73.89 & \resulttok{15,640} & \bestacc{71.94} & \resulttok{17,241}
 & 71.94 & \resulttok{16,158} & 46.11 & \resulttok{21,205} & \bestacc{65.97} & \resulttok{17,561} \\
30 & 100 & 72.78 & \resulttok{18,004} & 68.61 & \resulttok{19,436}
 & 72.50 & \resulttok{18,544} & \bestacc{47.22} & \resulttok{23,358} & 65.28 & \resulttok{19,836} \\
32.5 & 50 & \bestacc{75.56} & \resulttok{16,524} & 63.89 & \resulttok{18,977}
 & 68.33 & \resulttok{17,914} & 45.83 & \resulttok{22,424} & 63.40 & \resulttok{18,960} \\
50 & 100 & 74.44 & \resulttok{16,874} & 68.33 & \resulttok{19,020}
 & 70.00 & \resulttok{18,369} & 46.39 & \resulttok{22,932} & 64.79 & \resulttok{19,299} \\
500 & 100 & 74.17 & \resulttok{15,039} & 63.61 & \resulttok{17,293}
 & 71.94 & \resulttok{15,412} & 43.89 & \resulttok{20,952} & 63.40 & \resulttok{17,174} \\
\midrule
\rowcolor{metablue!7}
\multicolumn{12}{@{}l}{\textcolor{metablue}{\sffamily\bfseries Qwen3--4B}} \\
25 & 50 & 65.83 & \resulttok{18,888} & 58.33 & \resulttok{20,431}
 & 67.78 & \resulttok{18,758} & 43.33 & \resulttok{23,574} & 58.82 & \resulttok{20,413} \\
30 & 50 & 67.22 & \resulttok{18,254} & 56.94 & \resulttok{20,374}
 & 68.33 & \resulttok{18,132} & 39.72 & \resulttok{22,597} & 58.06 & \resulttok{19,839} \\
32.5 & 50 & 65.28 & \resulttok{18,990} & 58.33 & \resulttok{20,562}
 & 69.17 & \resulttok{18,513} & 42.22 & \resulttok{23,556} & 58.75 & \resulttok{20,405} \\
\rowcolor{metateal!5}
\textbf{35} & \textbf{200} & \bestacc{74.17} & \resulttok{15,273}
 & \bestacc{59.44} & \resulttok{17,466} & \bestacc{69.72} & \resulttok{16,181}
 & \bestacc{44.17} & \resulttok{20,143} & \bestacc{61.88} & \resulttok{17,265} \\
37.5 & 50 & 68.06 & \resulttok{17,344} & 57.22 & \resulttok{18,789}
 & 67.22 & \resulttok{17,437} & 38.89 & \resulttok{21,279} & 57.85 & \resulttok{18,712} \\
50 & 200 & 66.11 & \resulttok{16,556} & 57.78 & \resulttok{17,735}
 & 64.44 & \resulttok{16,318} & 39.17 & \resulttok{21,490} & 56.88 & \resulttok{18,025} \\
\bottomrule
\end{tabular*}
\end{table}

The highlighted rows in \cref{tab:srcl-coefficient} attain the highest observed Average for 8B
(65.97\%) and 4B (61.88\%).
The coefficient settings use separate stochastic trajectories, including independent seed-42
trajectories for the highlighted 4B run and several 8B refinements. Small differences therefore do
not establish a monotonic response or a general optimum.
}

\newcommand{\eospenaltyresults}{%
\paragraph{Results.}
\label{sec:late-eos-results}

\begin{table}[H]
\centering
\caption{Task-level EOS-Penalty results under non-thinking, 32K, Average@12 evaluation. Accuracy
(Acc., \%) and mean generated tokens (Tok.) are reported separately.}
\label{tab:eos-selected}
\footnotesize
\setlength{\tabcolsep}{2.0pt}
\renewcommand{\arraystretch}{1.04}
\begin{tabular*}{\textwidth}{@{\extracolsep{\fill}}lr*{5}{rr}@{}}
\toprule
\textbf{Model} & \textbf{Step}
 & \multicolumn{2}{c}{\textbf{AIME24}} & \multicolumn{2}{c}{\textbf{AIME25}}
 & \multicolumn{2}{c}{\textbf{AIME26}} & \multicolumn{2}{c}{\textbf{HMMT25}}
 & \multicolumn{2}{c}{\textbf{Average}} \\
\cmidrule(lr){3-4}\cmidrule(lr){5-6}\cmidrule(lr){7-8}\cmidrule(lr){9-10}\cmidrule(l){11-12}
 & & Acc. & Tok. & Acc. & Tok. & Acc. & Tok. & Acc. & Tok. & Acc. & Tok. \\
\midrule
Qwen3--8B & 200 & 77.50 & \resulttok{13,688} & 66.11 & \resulttok{16,052}
 & 63.06 & \resulttok{15,059} & 37.78 & \resulttok{19,469} & 61.11 & \resulttok{16,067} \\
Qwen3--4B & 150 & 42.78 & \resulttok{8,160} & 32.78 & \resulttok{6,846}
 & 35.00 & \resulttok{6,895} & 20.28 & \resulttok{6,774} & 32.71 & \resulttok{7,169} \\
\bottomrule
\end{tabular*}
\end{table}

\begin{table}[H]
\centering
\caption{Average@12 accuracy and mean output length across EOS-Penalty checkpoints.}
\label{tab:eos-curves}
\small
\setlength{\tabcolsep}{6pt}
\renewcommand{\arraystretch}{1.02}
\begin{tabular}{lrrrr}
\toprule
\textbf{Step} & \multicolumn{2}{c}{\textbf{Qwen3--8B}} & \multicolumn{2}{c}{\textbf{Qwen3--4B}} \\
\cmidrule(lr){2-3}\cmidrule(l){4-5}
 & Acc. & Tok. & Acc. & Tok. \\
\midrule
s50  & 44.93 & \resulttok{9,464}  & 29.51 & \resulttok{6,042} \\
s100 & 58.19 & \resulttok{14,273} & 29.51 & \resulttok{6,185} \\
s150 & 60.14 & \resulttok{15,631} & 32.71 & \resulttok{7,169} \\
s200 & 61.11 & \resulttok{16,067} & 32.57 & \resulttok{6,927} \\
\bottomrule
\end{tabular}
\end{table}

At 8B, EOS Penalty is 2,979 tokens shorter than DCE but 4.65 points less accurate and remains below
DCE+SRCL. At 4B, it reaches only 32.71\% with 7,169 tokens. Direct termination pressure therefore
shortens responses at a substantially larger accuracy cost than learning from concise verified
rewrites.
}

\FloatBarrier
\inferenceanalyses

\section{Conclusion}

We presented Dynamic Co-Evolution (DCE), which makes privileged self-distillation recursive: every
updated checkpoint becomes the next student and the next detached teacher, so revision learned in
one round shapes the next round's supervision. Self-Refined Concise Learning (SRCL) complements it
with shorter, answer-verified rewrites of the model's own responses; together, they deliver stronger and more token-efficient
recursive self-improvement across model scales.

\bibliographystyle{assets/plainnat}
\bibliography{paper}

\begin{thebibliography}{40}
\providecommand{\natexlab}[1]{#1}
\providecommand{\url}[1]{\texttt{#1}}
\expandafter\ifx\csname urlstyle\endcsname\relax
  \providecommand{\doi}[1]{doi: #1}\else
  \providecommand{\doi}{doi: \begingroup \urlstyle{rm}\Url}\fi

\bibitem[Agarwal et~al.(2024)Agarwal, Vieillard, Zhou, Stanczyk, Garea, Geist,
  and Bachem]{agarwal2024policy}
Rishabh Agarwal, Nino Vieillard, Yongchao Zhou, Piotr Stanczyk, Sabela~Ramos
  Garea, Matthieu Geist, and Olivier Bachem.
\newblock On-policy distillation of language models: Learning from
  self-generated mistakes.
\newblock In \emph{The Twelfth International Conference on Learning
  Representations, {ICLR} 2024, Vienna, Austria, May 7-11, 2024}.
  OpenReview.net, 2024.
\newblock \url{https://openreview.net/forum?id=3zKtaqxLhW}.

\bibitem[Brown et~al.(2024)Brown, Juravsky, Ehrlich, Clark, Le, R{\'{e}}, and
  Mirhoseini]{brown2024large}
Bradley C.~A. Brown, Jordan Juravsky, Ryan Ehrlich, Ronald Clark, Quoc~V. Le,
  Christopher R{\'{e}}, and Azalia Mirhoseini.
\newblock Large language monkeys: Scaling inference compute with repeated
  sampling.
\newblock \emph{CoRR}, abs/2407.21787, 2024.
\newblock \doi{10.48550/ARXIV.2407.21787}.
\newblock \url{https://doi.org/10.48550/arXiv.2407.21787}.

\bibitem[Damani et~al.(2025)Damani, Shenfeld, Peng, Bobu, and
  Andreas]{damani2024learning}
Mehul Damani, Idan Shenfeld, Andi Peng, Andreea Bobu, and Jacob Andreas.
\newblock Learning how hard to think: Input-adaptive allocation of {LM}
  computation.
\newblock In \emph{The Thirteenth International Conference on Learning
  Representations, {ICLR} 2025, Singapore, April 24-28, 2025}. OpenReview.net,
  2025.
\newblock \url{https://openreview.net/forum?id=6qUUgw9bAZ}.

\bibitem[Firooz et~al.(2025)Firooz, Liu, Lu, Hou, Xiong, Zhang, Jian, Zhu, Ma,
  Tao, et~al.]{firooz2025scaling}
Hamed Firooz, Rui Liu, Yuchen Lu, Zhenyu Hou, Fangzhou Xiong, Xiaoyang Zhang,
  Changshu Jian, Zhicheng Zhu, Jiayuan Ma, Jacob Tao, et~al.
\newblock Scaling reinforcement learning for content moderation with large
  language models.
\newblock \emph{CoRR}, abs/2512.20061, 2025.
\newblock \doi{10.48550/ARXIV.2512.20061}.
\newblock \url{https://doi.org/10.48550/arXiv.2512.20061}.

\bibitem[Gandhi et~al.(2025)Gandhi, Chakravarthy, Singh, Lile, and
  Goodman]{gandhi2025cognitive}
Kanishk Gandhi, Ayush Chakravarthy, Anikait Singh, Nathan Lile, and Noah~D.
  Goodman.
\newblock Cognitive behaviors that enable self-improving reasoners, or, four
  habits of highly effective stars.
\newblock \emph{CoRR}, abs/2503.01307, 2025.
\newblock \doi{10.48550/ARXIV.2503.01307}.
\newblock \url{https://doi.org/10.48550/arXiv.2503.01307}.

\bibitem[Ghosal et~al.(2025)Ghosal, Chakraborty, Reddy, Lu, Wang, Manocha,
  Huang, Ghavamzadeh, and Bedi]{ghosal2025thinkingmore}
Soumya~Suvra Ghosal, Souradip Chakraborty, Avinash Reddy, Yifu Lu, Mengdi Wang,
  Dinesh Manocha, Furong Huang, Mohammad Ghavamzadeh, and Amrit~Singh Bedi.
\newblock Does thinking more always help? mirage of test-time scaling in
  reasoning models.
\newblock In Danielle Belgrave, Cheng Zhang, Laura~N. Montoya, Hsuan{-}Tien
  Lin, Razvan Pascanu, Piotr Koniusz, Marzyeh Ghassemi, Nancy Chen, Iv{\'{a}}n
  Vladimir~Meza Ru{\'{\i}}z, and Arturo Loaiza{-}Bonilla, editors,
  \emph{Advances in Neural Information Processing Systems 38: Annual Conference
  on Neural Information Processing Systems 2025, NeurIPS 2025, San Diego, CA,
  USA, December 2-7, 2025 / Mexico City, Mexico, November 30 - December 5,
  2025}, 2025.
\newblock
  \url{http://papers.nips.cc/paper\_files/paper/2025/hash/fc067ac218430c409d6f65403328f740-Abstract-Conference.html}.

\bibitem[Guha et~al.(2025)Guha, Marten, Keh, Raoof, Smyrnis, Bansal, Nezhurina,
  Mercat, Vu, Sprague, et~al.]{guha2025openthoughtsdatarecipesreasoning}
Etash~Kumar Guha, Ryan Marten, Sedrick Keh, Negin Raoof, Georgios Smyrnis,
  Hritik Bansal, Marianna Nezhurina, Jean Mercat, Trung Vu, Zayne Sprague,
  et~al.
\newblock Openthoughts: Data recipes for reasoning models.
\newblock \emph{CoRR}, abs/2506.04178, 2025.
\newblock \doi{10.48550/ARXIV.2506.04178}.
\newblock \url{https://doi.org/10.48550/arXiv.2506.04178}.

\bibitem[G{\"{u}}l{\c{c}}ehre et~al.(2023)G{\"{u}}l{\c{c}}ehre, Paine,
  Srinivasan, Konyushkova, Weerts, Sharma, Siddhant, Ahern, Wang, Gu,
  et~al.]{gulcehre2023reinforced}
{\c{C}}aglar G{\"{u}}l{\c{c}}ehre, Tom~Le Paine, Srivatsan Srinivasan, Ksenia
  Konyushkova, Lotte Weerts, Abhishek Sharma, Aditya Siddhant, Alex Ahern,
  Miaosen Wang, Chenjie Gu, et~al.
\newblock Reinforced self-training (rest) for language modeling.
\newblock \emph{CoRR}, abs/2308.08998, 2023.
\newblock \doi{10.48550/ARXIV.2308.08998}.
\newblock \url{https://doi.org/10.48550/arXiv.2308.08998}.

\bibitem[Guo et~al.(2025)Guo, Yang, Zhang, Song, Wang, Zhu, Xu, Zhang, Ma, Bi,
  et~al.]{guo2025deepseek}
Daya Guo, Dejian Yang, Haowei Zhang, Junxiao Song, Peiyi Wang, Qihao Zhu,
  Runxin Xu, Ruoyu Zhang, Shirong Ma, Xiao Bi, et~al.
\newblock Deepseek-r1 incentivizes reasoning in llms through reinforcement
  learning.
\newblock \emph{Nat.}, 645\penalty0 (8081):\penalty0 633--638, 2025.
\newblock \doi{10.1038/S41586-025-09422-Z}.
\newblock \url{https://doi.org/10.1038/s41586-025-09422-z}.

\bibitem[{Harvard--MIT Mathematics Tournament}(2026)]{hmmtArchive}
{Harvard--MIT Mathematics Tournament}.
\newblock Past tournaments, 2026.
\newblock \url{https://www.hmmt.org/www/archive/problems}.
\newblock Accessed August 26, 2026.

\bibitem[Hendrycks et~al.(2021)Hendrycks, Burns, Kadavath, Arora, Basart, Tang,
  Song, and Steinhardt]{hendrycks2021math}
Dan Hendrycks, Collin Burns, Saurav Kadavath, Akul Arora, Steven Basart, Eric
  Tang, Dawn Song, and Jacob Steinhardt.
\newblock Measuring mathematical problem solving with the {MATH} dataset.
\newblock In Joaquin Vanschoren and Sai{-}Kit Yeung, editors, \emph{Proceedings
  of the Neural Information Processing Systems Track on Datasets and Benchmarks
  1, NeurIPS Datasets and Benchmarks 2021, December 2021, virtual}, 2021.
\newblock
  \url{https://datasets-benchmarks-proceedings.neurips.cc/paper/2021/hash/be83ab3ecd0db773eb2dc1b0a17836a1-Abstract-round2.html}.

\bibitem[Hu et~al.(2022)Hu, Shen, Wallis, Allen{-}Zhu, Li, Wang, Wang, and
  Chen]{hu2022lora}
Edward~J. Hu, Yelong Shen, Phillip Wallis, Zeyuan Allen{-}Zhu, Yuanzhi Li,
  Shean Wang, Lu~Wang, and Weizhu Chen.
\newblock Lora: Low-rank adaptation of large language models.
\newblock In \emph{The Tenth International Conference on Learning
  Representations, {ICLR} 2022, Virtual Event, April 25-29, 2022}.
  OpenReview.net, 2022.
\newblock \url{https://openreview.net/forum?id=nZeVKeeFYf9}.

\bibitem[Huang et~al.(2024)Huang, Chen, Mishra, Zheng, Yu, Song, and
  Zhou]{huang2023cannot}
Jie Huang, Xinyun Chen, Swaroop Mishra, Huaixiu~Steven Zheng, Adams~Wei Yu,
  Xinying Song, and Denny Zhou.
\newblock Large language models cannot self-correct reasoning yet.
\newblock In \emph{The Twelfth International Conference on Learning
  Representations, {ICLR} 2024, Vienna, Austria, May 7-11, 2024}.
  OpenReview.net, 2024.
\newblock \url{https://openreview.net/forum?id=IkmD3fKBPQ}.

\bibitem[Lee et~al.(2025)Lee, Sun, Wendler, Vi{\'{e}}gas, and
  Wattenberg]{lee2025geometry}
Andrew Lee, Lihao Sun, Chris Wendler, Fernanda~B. Vi{\'{e}}gas, and Martin
  Wattenberg.
\newblock The geometry of self-verification in a task-specific reasoning model.
\newblock \emph{CoRR}, abs/2504.14379, 2025.
\newblock \doi{10.48550/ARXIV.2504.14379}.
\newblock \url{https://doi.org/10.48550/arXiv.2504.14379}.

\bibitem[Lightman et~al.(2024)Lightman, Kosaraju, Burda, Edwards, Baker, Lee,
  Leike, Schulman, Sutskever, and Cobbe]{lightman2023let}
Hunter Lightman, Vineet Kosaraju, Yuri Burda, Harrison Edwards, Bowen Baker,
  Teddy Lee, Jan Leike, John Schulman, Ilya Sutskever, and Karl Cobbe.
\newblock Let's verify step by step.
\newblock In \emph{The Twelfth International Conference on Learning
  Representations, {ICLR} 2024, Vienna, Austria, May 7-11, 2024}.
  OpenReview.net, 2024.
\newblock \url{https://openreview.net/forum?id=v8L0pN6EOi}.

\bibitem[Liu et~al.(2025)Liu, Chen, Li, Qi, Pang, Du, Lee, and
  Lin]{liu2025understanding}
Zichen Liu, Changyu Chen, Wenjun Li, Penghui Qi, Tianyu Pang, Chao Du, Wee~Sun
  Lee, and Min Lin.
\newblock Understanding r1-zero-like training: {A} critical perspective.
\newblock \emph{CoRR}, abs/2503.20783, 2025.
\newblock \doi{10.48550/ARXIV.2503.20783}.
\newblock \url{https://doi.org/10.48550/arXiv.2503.20783}.

\bibitem[Loshchilov and Hutter(2019)]{loshchilov2017decoupled}
Ilya Loshchilov and Frank Hutter.
\newblock Decoupled weight decay regularization.
\newblock In \emph{7th International Conference on Learning Representations,
  {ICLR} 2019, New Orleans, LA, USA, May 6-9, 2019}. OpenReview.net, 2019.
\newblock \url{https://openreview.net/forum?id=Bkg6RiCqY7}.

\bibitem[Lu and Lab(2025)]{lu2025onpolicydistillation}
Kevin Lu and Thinking~Machines Lab.
\newblock On-policy distillation.
\newblock \emph{Thinking Machines Lab: Connectionism}, 2025.
\newblock \doi{10.64434/tml.20251026}.
\newblock https://thinkingmachines.ai/blog/on-policy-distillation.

\bibitem[{Mathematical Association of America}(2026)]{maaAIME}
{Mathematical Association of America}.
\newblock {MAA} invitational competitions, 2026.
\newblock \url{https://maa.org/maa-invitational-competitions/}.
\newblock Accessed August 26, 2026.

\bibitem[Muennighoff et~al.(2025)Muennighoff, Yang, Shi, Li, Fei{-}Fei,
  Hajishirzi, Zettlemoyer, Liang, Cand{\`{e}}s, and
  Hashimoto]{muennighoff2025s1}
Niklas Muennighoff, Zitong Yang, Weijia Shi, Xiang~Lisa Li, Li~Fei{-}Fei,
  Hannaneh Hajishirzi, Luke Zettlemoyer, Percy Liang, Emmanuel~J. Cand{\`{e}}s,
  and Tatsunori Hashimoto.
\newblock s1: Simple test-time scaling.
\newblock In Christos Christodoulopoulos, Tanmoy Chakraborty, Carolyn Rose, and
  Violet Peng, editors, \emph{Proceedings of the 2025 Conference on Empirical
  Methods in Natural Language Processing, {EMNLP} 2025, Suzhou, China, November
  4-9, 2025}, pages 20275--20321. Association for Computational Linguistics,
  2025.
\newblock \doi{10.18653/V1/2025.EMNLP-MAIN.1025}.
\newblock \url{https://doi.org/10.18653/v1/2025.emnlp-main.1025}.

\bibitem[Pan et~al.(2026)Pan, Tao, Zhai, Zhang, Liu, Ding, Liu, and
  Wen]{pan2026rlcsd}
Leyi Pan, Shuchang Tao, Yunpeng Zhai, Lingzhe Zhang, Zhaoyang Liu, Bolin Ding,
  Aiwei Liu, and Lijie Wen.
\newblock {RLCSD:} reinforcement learning with contrastive on-policy
  self-distillation.
\newblock \emph{CoRR}, abs/2606.11709, 2026.
\newblock \doi{10.48550/ARXIV.2606.11709}.
\newblock \url{https://doi.org/10.48550/arXiv.2606.11709}.

\bibitem[{Qwen Team}(2025)]{qwen3}
{Qwen Team}.
\newblock Qwen3 technical report.
\newblock \emph{CoRR}, abs/2505.09388, 2025.
\newblock \doi{10.48550/ARXIV.2505.09388}.
\newblock \url{https://doi.org/10.48550/arXiv.2505.09388}.

\bibitem[Rein et~al.(2023)Rein, Hou, Stickland, Petty, Pang, Dirani, Michael,
  and Bowman]{rein2023gpqa}
David Rein, Betty~Li Hou, Asa~Cooper Stickland, Jackson Petty, Richard~Yuanzhe
  Pang, Julien Dirani, Julian Michael, and Samuel~R. Bowman.
\newblock {GPQA:} {A} graduate-level google-proof q{\&}a benchmark.
\newblock \emph{CoRR}, abs/2311.12022, 2023.
\newblock \doi{10.48550/ARXIV.2311.12022}.
\newblock \url{https://doi.org/10.48550/arXiv.2311.12022}.

\bibitem[Shao et~al.(2024)Shao, Wang, Zhu, Xu, Song, Bi, Zhang, Zhang, Li, Wu,
  et~al.]{shao2024deepseekmath}
Zhihong Shao, Peiyi Wang, Qihao Zhu, Runxin Xu, Junxiao Song, Xiao Bi, Haowei
  Zhang, Mingchuan Zhang, Y.~K. Li, Y.~Wu, et~al.
\newblock Deepseekmath: Pushing the limits of mathematical reasoning in open
  language models.
\newblock \emph{CoRR}, abs/2402.03300, 2024.
\newblock \doi{10.48550/ARXIV.2402.03300}.
\newblock \url{https://doi.org/10.48550/arXiv.2402.03300}.

\bibitem[Shinn et~al.(2023)Shinn, Cassano, Gopinath, Narasimhan, and
  Yao]{shinn2023reflexion}
Noah Shinn, Federico Cassano, Ashwin Gopinath, Karthik Narasimhan, and Shunyu
  Yao.
\newblock Reflexion: language agents with verbal reinforcement learning.
\newblock In Alice Oh, Tristan Naumann, Amir Globerson, Kate Saenko, Moritz
  Hardt, and Sergey Levine, editors, \emph{Advances in Neural Information
  Processing Systems 36: Annual Conference on Neural Information Processing
  Systems 2023, NeurIPS 2023, New Orleans, LA, USA, December 10 - 16, 2023},
  2023.
\newblock
  \url{http://papers.nips.cc/paper\_files/paper/2023/hash/1b44b878bb782e6954cd888628510e90-Abstract-Conference.html}.

\bibitem[Snell et~al.(2025)Snell, Lee, Xu, and Kumar]{snell2024scaling}
Charlie~Victor Snell, Jaehoon Lee, Kelvin Xu, and Aviral Kumar.
\newblock Scaling {LLM} test-time compute optimally can be more effective than
  scaling parameters for reasoning.
\newblock In \emph{The Thirteenth International Conference on Learning
  Representations, {ICLR} 2025, Singapore, April 24-28, 2025}. OpenReview.net,
  2025.
\newblock \url{https://openreview.net/forum?id=4FWAwZtd2n}.

\bibitem[Wang et~al.(2025)Wang, Feng, Chen, Chu, Krishna, and
  Zhou]{wang2025wait}
Chenlong Wang, Yuanning Feng, Dongping Chen, Zhaoyang Chu, Ranjay Krishna, and
  Tianyi Zhou.
\newblock Wait, we don't need to "wait"! removing thinking tokens improves
  reasoning efficiency.
\newblock In Christos Christodoulopoulos, Tanmoy Chakraborty, Carolyn Rose, and
  Violet Peng, editors, \emph{Findings of the Association for Computational
  Linguistics: {EMNLP} 2025, Suzhou, China, November 4-9, 2025}, pages
  7459--7482. Association for Computational Linguistics, 2025.
\newblock \doi{10.18653/V1/2025.FINDINGS-EMNLP.394}.
\newblock \url{https://doi.org/10.18653/v1/2025.findings-emnlp.394}.

\bibitem[Wu et~al.(2025)Wu, Sun, Li, Welleck, and Yang]{wu2025inference}
Yangzhen Wu, Zhiqing Sun, Shanda Li, Sean Welleck, and Yiming Yang.
\newblock Inference scaling laws: An empirical analysis of compute-optimal
  inference for {LLM} problem-solving.
\newblock In \emph{The Thirteenth International Conference on Learning
  Representations, {ICLR} 2025, Singapore, April 24-28, 2025}. OpenReview.net,
  2025.
\newblock \url{https://openreview.net/forum?id=VNckp7JEHn}.

\bibitem[Yang et~al.(2026)Yang, Qin, Si, Chen, Gu, Yao, Lin, Wang, Wang, and
  Duan]{yang2026rlsd}
Chenxu Yang, Chuanyu Qin, Qingyi Si, Minghui Chen, Naibin Gu, Dingyu Yao, Zheng
  Lin, Weiping Wang, Jiaqi Wang, and Nan Duan.
\newblock Self-distilled {RLVR}.
\newblock \emph{CoRR}, abs/2604.03128, 2026.
\newblock \doi{10.48550/ARXIV.2604.03128}.
\newblock \url{https://doi.org/10.48550/arXiv.2604.03128}.

\bibitem[Yin and Shi(2026)]{yin-shi-2026-individual}
Shangjian Yin and Zhouxing Shi.
\newblock From individual to common: An early exploration of consensus in
  non-verifiable data for balanced preference optimization.
\newblock In \emph{Proceedings of the 64th Annual Meeting of the Association
  for Computational Linguistics (Volume 1: Long Papers)}, pages 34612--34630.
  Association for Computational Linguistics, 2026.
\newblock \doi{10.18653/v1/2026.acl-long.1598}.
\newblock \url{https://aclanthology.org/2026.acl-long.1598/}.

\bibitem[Yin et~al.(2025)Yin, Liang, Ding, Qian, Shi, Li, and Xie]{yin2025pika}
Shangjian Yin, Shining Liang, Wenbiao Ding, Yuli Qian, Zhouxing Shi, Hongzhi
  Li, and Yutao Xie.
\newblock {PIKA}: Expert-level synthetic datasets for post-training alignment
  from scratch.
\newblock \emph{CoRR}, abs/2510.06670, 2025.
\newblock \doi{10.48550/ARXIV.2510.06670}.
\newblock \url{https://arxiv.org/abs/2510.06670}.

\bibitem[Yin et~al.(2026{\natexlab{a}})Yin, Fu, Dong, and Shi]{yin2026grlo}
Shangjian Yin, Yu~Fu, Yue Dong, and Zhouxing Shi.
\newblock {GRLO}: Towards generalizable reinforcement learning in open-ended
  environments from zero.
\newblock \emph{CoRR}, abs/2605.15464, 2026{\natexlab{a}}.
\newblock \doi{10.48550/ARXIV.2605.15464}.
\newblock \url{https://arxiv.org/abs/2605.15464}.

\bibitem[Yin et~al.(2026{\natexlab{b}})Yin, Wei, Zhu, Chen, and
  Meng]{yin-etal-2026-aligning}
Shangjian Yin, Zhepei Wei, Xinyu Zhu, Wei{-}Lin Chen, and Yu~Meng.
\newblock Aligning large language models via fully self-synthetic data.
\newblock In \emph{Proceedings of the 64th Annual Meeting of the Association
  for Computational Linguistics (Volume 1: Long Papers)}, pages 34553--34568.
  Association for Computational Linguistics, 2026{\natexlab{b}}.
\newblock \doi{10.18653/v1/2026.acl-long.1595}.
\newblock \url{https://aclanthology.org/2026.acl-long.1595/}.

\bibitem[Yu et~al.(2025)Yu, Zhang, Zhu, Yuan, Zuo, Yue, Dai, Fan, Liu, Liu,
  et~al.]{yu2025dapo}
Qiying Yu, Zheng Zhang, Ruofei Zhu, Yufeng Yuan, Xiaochen Zuo, Yu~Yue, Weinan
  Dai, Tiantian Fan, Gaohong Liu, Juncai Liu, et~al.
\newblock {DAPO:} an open-source {LLM} reinforcement learning system at scale.
\newblock In Danielle Belgrave, Cheng Zhang, Laura~N. Montoya, Hsuan{-}Tien
  Lin, Razvan Pascanu, Piotr Koniusz, Marzyeh Ghassemi, Nancy Chen, Iv{\'{a}}n
  Vladimir~Meza Ru{\'{\i}}z, and Arturo Loaiza{-}Bonilla, editors,
  \emph{Advances in Neural Information Processing Systems 38: Annual Conference
  on Neural Information Processing Systems 2025, NeurIPS 2025, San Diego, CA,
  USA, December 2-7, 2025 / Mexico City, Mexico, November 30 - December 5,
  2025}, 2025.
\newblock
  \url{http://papers.nips.cc/paper\_files/paper/2025/hash/a4277440d50f1f15d2cb4c14f7e0c0d2-Abstract-Conference.html}.

\bibitem[Zelikman et~al.(2022)Zelikman, Wu, Mu, and Goodman]{zelikman2022star}
Eric Zelikman, Yuhuai Wu, Jesse Mu, and Noah~D. Goodman.
\newblock Star: Bootstrapping reasoning with reasoning.
\newblock In Sanmi Koyejo, S.~Mohamed, A.~Agarwal, Danielle Belgrave, K.~Cho,
  and A.~Oh, editors, \emph{Advances in Neural Information Processing Systems
  35: Annual Conference on Neural Information Processing Systems 2022, NeurIPS
  2022, New Orleans, LA, USA, November 28 - December 9, 2022}, 2022.
\newblock
  \url{http://papers.nips.cc/paper\_files/paper/2022/hash/639a9a172c044fbb64175b5fad42e9a5-Abstract-Conference.html}.

\bibitem[Zeng et~al.(2025)Zeng, Huang, Liu, Liu, He, Ma, and
  He]{zeng2025simplerl}
Weihao Zeng, Yuzhen Huang, Qian Liu, Wei Liu, Keqing He, Zejun Ma, and Junxian
  He.
\newblock Simplerl-zoo: Investigating and taming zero reinforcement learning
  for open base models in the wild.
\newblock \emph{CoRR}, abs/2503.18892, 2025.
\newblock \doi{10.48550/ARXIV.2503.18892}.
\newblock \url{https://doi.org/10.48550/arXiv.2503.18892}.

\bibitem[Zhang et~al.(2025{\natexlab{a}})Zhang, Chen, Pan, Zhao, Panda, Li, and
  He]{zhang2025reasoningright}
Anqi Zhang, Yulin Chen, Jane Pan, Chen Zhao, Aurojit Panda, Jinyang Li, and
  He~He.
\newblock Reasoning models know when they're right: Probing hidden states for
  self-verification.
\newblock \emph{CoRR}, abs/2504.05419, 2025{\natexlab{a}}.
\newblock \doi{10.48550/ARXIV.2504.05419}.
\newblock \url{https://doi.org/10.48550/arXiv.2504.05419}.

\bibitem[Zhang et~al.(2025{\natexlab{b}})Zhang, Zheng, Wu, Zhang, Lin, Yu, Liu,
  Zhou, and Lin]{zhang2025lessons}
Zhenru Zhang, Chujie Zheng, Yangzhen Wu, Beichen Zhang, Runji Lin, Bowen Yu,
  Dayiheng Liu, Jingren Zhou, and Junyang Lin.
\newblock The lessons of developing process reward models in mathematical
  reasoning.
\newblock In Wanxiang Che, Joyce Nabende, Ekaterina Shutova, and Mohammad~Taher
  Pilehvar, editors, \emph{Findings of the Association for Computational
  Linguistics, {ACL} 2025, Vienna, Austria, July 27 - August 1, 2025}, volume
  {ACL} 2025 of \emph{Findings of {ACL}}, pages 10495--10516. Association for
  Computational Linguistics, 2025{\natexlab{b}}.
\newblock \doi{10.18653/V1/2025.FINDINGS-ACL.547}.
\newblock \url{https://doi.org/10.18653/v1/2025.findings-acl.547}.

\bibitem[Zhao et~al.(2026)Zhao, Xie, Liu, Huang, Pang, Chen, and
  Grover]{zhao2026selfdistilled}
Siyan Zhao, Zhihui Xie, Mengchen Liu, Jing Huang, Guan Pang, Feiyu Chen, and
  Aditya Grover.
\newblock Self-distilled reasoner: On-policy self-distillation for large
  language models.
\newblock \emph{CoRR}, abs/2601.18734, 2026.
\newblock \doi{10.48550/ARXIV.2601.18734}.
\newblock \url{https://doi.org/10.48550/arXiv.2601.18734}.

\bibitem[Zhu et~al.(2025)Zhu, Jiang, Khalili, and Zhu]{zhu2025emergence}
Xudong Zhu, Jiachen Jiang, Mohammad~Mahdi Khalili, and Zhihui Zhu.
\newblock From emergence to control: Probing and modulating self-reflection in
  language models.
\newblock \emph{CoRR}, abs/2506.12217, 2025.
\newblock \doi{10.48550/ARXIV.2506.12217}.
\newblock \url{https://doi.org/10.48550/arXiv.2506.12217}.

\end{thebibliography}

\appendix

\section{Does Recursive Improvement Scale to a Larger Model?}
\label{sec:qwen14}

\begin{table}[H]
\centering
\caption{Recursive self-improvement on Qwen3-14B. Results use non-thinking, 32K, Average@12
evaluation; Acc. denotes accuracy (\%) and Tok. mean generated tokens.}
\label{tab:qwen14-trajectory}
\footnotesize
\setlength{\tabcolsep}{1.8pt}
\renewcommand{\arraystretch}{1.08}
\begin{tabular*}{0.99\textwidth}{@{\extracolsep{\fill}}l*{5}{r@{\hspace{1.8pt}}r}@{}}
\toprule
\textbf{Step}
 & \multicolumn{2}{c}{\textbf{AIME24}}
 & \multicolumn{2}{c}{\textbf{AIME25}}
 & \multicolumn{2}{c}{\textbf{AIME26}}
 & \multicolumn{2}{c}{\textbf{HMMT25}}
 & \multicolumn{2}{c}{\textcolor{metablue}{\textbf{Average}}} \\
\cmidrule(lr){2-3}\cmidrule(lr){4-5}\cmidrule(lr){6-7}\cmidrule(lr){8-9}\cmidrule(l){10-11}
 & Acc. & Tok. & Acc. & Tok. & Acc. & Tok. & Acc. & Tok.
 & \textcolor{metablue}{Acc.} & \textcolor{metablue}{Tok.} \\
\midrule
Base & 30.56 & \resulttok{4,056} & 24.17 & \resulttok{3,094}
 & 19.44 & \resulttok{4,045} & 11.39 & \resulttok{2,998} & 21.39 & \resulttok{3,548} \\
10 & 36.39 & \resulttok{5,044} & 25.83 & \resulttok{4,589}
 & 26.39 & \resulttok{4,604} & 15.00 & \resulttok{4,102} & 25.90 & \resulttok{4,585} \\
20 & 55.28 & \resulttok{7,377} & 42.50 & \resulttok{7,255}
 & 45.28 & \resulttok{7,717} & 27.50 & \resulttok{7,502} & 42.64 & \resulttok{7,463} \\
30 & \bestacc{76.11} & \resulttok{14,391} & \bestacc{71.94} & \resulttok{15,973}
 & 69.72 & \resulttok{14,539} & 47.78 & \resulttok{18,605}
 & \bestacc{66.39} & \resulttok{15,877} \\
50 & 71.94 & \resulttok{18,111} & 69.72 & \resulttok{19,975}
 & \bestacc{70.28} & \resulttok{18,792} & \bestacc{48.89} & \resulttok{23,323}
 & 65.21 & \resulttok{20,050} \\
\bottomrule
\end{tabular*}
\end{table}

To explore whether recursive improvement extends to a larger model, we train Qwen3-14B with the
same DCE+SRCL framework and track its early trajectory in \cref{tab:qwen14-trajectory}. Average
accuracy rises from 21.39\% at initialization to 42.64\% at step 20 and 66.39\% at step 30, a
total gain of 45.00 percentage points. The improvement spans all four benchmarks and is largest
on AIME26, which rises by 50.28 percentage points, from 19.44\% to 69.72\%. Step 50 retains a
comparable 65.21\%, although mean output grows from 15,877 to 20,050 tokens. These results show
strong round-to-round improvement at 14B while indicating that early stopping remains important
for preserving the accuracy--length frontier.

\section{Gemma-4-12B-IT Training Details and Additional Results}
\label{app:gemma12b-details}

The cross-family comparison appears in \cref{tab:gemma12b-transfer}. The exact model ID is
\texttt{google/gemma-4-12B-it}. The matched DCE runs use the same 14,717-problem OPSD shard,
assistant-side privileged conditioning, non-thinking student and teacher branches, and 200-step
LoRA training with seed 42 (rank 128, alpha 256, global batch 16, and learning rate
$2.5\times10^{-6}$). Their only objective-level difference is $\lambda_S=0$ for DCE and
$\lambda_S=25$ for DCE+SRCL, with $\lambda_G=5\times10^4$ in both.

Placement also matters on Gemma-4-12B-IT. Assistant-side conditioning reaches 63.61\%
Average@12, compared with 55.76\%
for user-side reference-last and 54.31\% for user-side instruction-last
(\cref{tab:gemma12b-context-placement}). Both user-side variants attain their best observed result
at step 10, whereas the assistant-side run continues improving through step 100.

\begin{table}[H]
\centering
\caption{Privileged-solution placement on Gemma-4-12B-IT under non-thinking, 32K, Average@12
evaluation. Assistant-side uses checkpoint 100; both user-side variants use checkpoint 10. Each
cell reports accuracy (\%) / mean generated tokens.}
\label{tab:gemma12b-context-placement}
\scriptsize
\setlength{\tabcolsep}{2.0pt}
\renewcommand{\arraystretch}{1.10}
\begin{tabular*}{\textwidth}{@{\extracolsep{\fill}}lccccc@{}}
\toprule
\textbf{GOLD placement} & \textbf{AIME24} & \textbf{AIME25} & \textbf{AIME26}
& \textbf{HMMT25} & \textcolor{metablue}{\textbf{Average}} \\
\midrule
\rowcolor{metateal!5}
\textbf{Assistant-side}
& \bestacc{72.50} / 6,661 & \bestacc{64.72} / 8,341
& \bestacc{68.06} / 7,965 & \bestacc{49.17} / 11,137
& \bestacc{63.61} / 8,526 \\
User-side: reference-last
& 67.22 / 8,242 & 53.89 / 12,188 & 61.94 / 8,994 & 40.00 / 14,066
& 55.76 / 10,872 \\
User-side: instruction-last
& 65.00 / 7,692 & 52.50 / 10,646 & 60.28 / 9,169 & 39.44 / 12,462
& 54.31 / 9,992 \\
\bottomrule
\end{tabular*}
\end{table}

\begin{table}[H]
\centering
\caption{Matched Gemma-4-12B-IT training trajectories. Accuracy is four-task Average@12 (\%),
Tok. is mean generated tokens, and $\Delta$ is DCE+SRCL minus DCE accuracy in percentage points.}
\label{tab:gemma12b-trajectory}
\footnotesize
\setlength{\tabcolsep}{4.0pt}
\renewcommand{\arraystretch}{1.04}
\begin{tabular*}{0.78\textwidth}{@{\extracolsep{\fill}}lrrrrr@{}}
\toprule
\textbf{Step}
 & \multicolumn{2}{c}{\textcolor{metablue}{\textbf{DCE}}}
 & \multicolumn{2}{c}{\textcolor{metateal}{\textbf{DCE+SRCL}}}
 & \textbf{$\Delta$ Acc.} \\
\cmidrule(lr){2-3}\cmidrule(lr){4-5}
 & Acc. & Tok. & Acc. & Tok. & (pp) \\
\midrule
10  & 55.76 & \resulttok{8,381}  & 56.39 & \resulttok{8,357}  & $+0.63$ \\
20  & 59.17 & \resulttok{7,866}  & 57.36 & \resulttok{8,111}  & $-1.81$ \\
30  & 59.79 & \resulttok{7,691}  & 59.79 & \resulttok{8,424}  & $\phantom{+}0.00$ \\
50  & 61.04 & \resulttok{7,701}  & 62.85 & \resulttok{7,328}  & $+1.81$ \\
\rowcolor{metateal!4}
100 & \bestacc{62.01} & \resulttok{8,812} & \bestacc{63.61} & \resulttok{8,526} & $+1.60$ \\
150 & 61.04 & \resulttok{10,015} & 62.08 & \resulttok{9,457} & $+1.04$ \\
200 & 60.76 & \resulttok{9,968}  & 61.46 & \resulttok{10,439} & $+0.69$ \\
\bottomrule
\end{tabular*}
\end{table}

\section{Complete Qwen3-8B and Qwen3-4B Trajectories}
\label{app:full-trajectories}

\Cref{tab:qwen8-appendix,tab:qwen4-appendix} report checkpoint-wise Average@12 accuracy and mean
output length for Qwen3-8B and Qwen3-4B, complementing the learning curves in
\cref{fig:learning-dynamics}. Missing entries denote unevaluated checkpoints. The GRPO columns show
the available trajectories, while \cref{tab:main-results} reports the tuned baseline comparison.

\begin{table}[H]
\centering
\caption{Qwen3-8B checkpoint trajectories. Adjacent columns report Average@12 accuracy
(Acc., \%) and mean generated tokens (Tok.); blue bold marks the highest accuracy for each method.
The base model obtains 19.51\% with 3,922 tokens.}
\label{tab:qwen8-appendix}
\footnotesize
\setlength{\tabcolsep}{3pt}
\renewcommand{\arraystretch}{1.02}
\begin{tabular*}{0.96\textwidth}{@{\extracolsep{\fill}}l*{4}{r@{\hspace{2.5pt}}r}@{}}
\toprule
\textbf{Step} & \multicolumn{2}{c}{\textbf{GRPO}} & \multicolumn{2}{c}{\textbf{OPSD}}
 & \multicolumn{2}{c}{\textcolor{metablue}{\textbf{DCE}}}
 & \multicolumn{2}{c}{\textcolor{metateal}{\textbf{+ SRCL}}} \\
\cmidrule(lr){2-3}\cmidrule(lr){4-5}\cmidrule(lr){6-7}\cmidrule(l){8-9}
 & Acc. & Tok. & Acc. & Tok. & Acc. & Tok. & Acc. & Tok. \\
\midrule
10  & 18.47 & \resulttok{3,860} & 18.89 & \resulttok{3,942}
 & 22.85 & \resulttok{5,031} & 23.54 & \resulttok{5,016} \\
20  & 19.72 & \resulttok{3,942} & 20.28 & \resulttok{4,684}
 & 32.36 & \resulttok{7,168} & 33.75 & \resulttok{7,064} \\
30  & 20.21 & \resulttok{3,914} & 21.32 & \resulttok{5,240}
 & 45.49 & \resulttok{11,141} & 46.67 & \resulttok{10,759} \\
50  & 19.51 & \resulttok{3,774} & 28.89 & \resulttok{6,580}
 & 64.44 & \resulttok{18,728} & 61.32 & \resulttok{15,959} \\
100 & 19.86 & \resulttok{3,841} & \bestacc{30.35} & \resulttok{6,009}
 & 64.93 & \resulttok{19,036} & \bestacc{65.97} & \resulttok{17,561} \\
150 & \bestacc{20.28} & \resulttok{3,705} & 29.65 & \resulttok{5,975}
 & \bestacc{65.76} & \resulttok{19,046} & 65.14 & \resulttok{18,057} \\
200 & 20.21 & \resulttok{3,827} & 28.06 & \resulttok{5,506}
 & 64.31 & \resulttok{19,072} & 64.58 & \resulttok{18,225} \\
\bottomrule
\end{tabular*}
\end{table}

\begin{table}[H]
\centering
\caption{Qwen3-4B checkpoint trajectories, using the same convention as
\cref{tab:qwen8-appendix}. The base model obtains 17.57\% with 3,805 tokens.}
\label{tab:qwen4-appendix}
\footnotesize
\setlength{\tabcolsep}{3pt}
\renewcommand{\arraystretch}{1.02}
\begin{tabular*}{0.96\textwidth}{@{\extracolsep{\fill}}l*{4}{r@{\hspace{2.5pt}}r}@{}}
\toprule
\textbf{Step} & \multicolumn{2}{c}{\textbf{GRPO}} & \multicolumn{2}{c}{\textbf{OPSD}}
 & \multicolumn{2}{c}{\textcolor{metablue}{\textbf{DCE}}}
 & \multicolumn{2}{c}{\textcolor{metateal}{\textbf{+ SRCL}}} \\
\cmidrule(lr){2-3}\cmidrule(lr){4-5}\cmidrule(lr){6-7}\cmidrule(l){8-9}
 & Acc. & Tok. & Acc. & Tok. & Acc. & Tok. & Acc. & Tok. \\
\midrule
10  & 17.08 & \resulttok{3,622} & 18.96 & \resulttok{4,078}
 & 22.99 & \resulttok{4,792} & 22.43 & \resulttok{4,912} \\
20  & \bestacc{18.33} & \resulttok{3,653} & 19.65 & \resulttok{4,726}
 & 30.00 & \resulttok{7,723} & 30.21 & \resulttok{7,388} \\
30  & 17.43 & \resulttok{3,677} & 20.49 & \resulttok{6,136}
 & 54.38 & \resulttok{16,975} & 53.82 & \resulttok{15,096} \\
50  & 17.99 & \resulttok{3,478} & 18.26 & \resulttok{7,676}
 & \bestacc{60.00} & \resulttok{19,360} & 57.99 & \resulttok{15,044} \\
100 & 17.99 & \resulttok{3,479} & \bestacc{22.85} & \resulttok{8,350}
 & 56.04 & \resulttok{22,423} & 60.83 & \resulttok{17,152} \\
150 & 17.99 & \resulttok{3,444} & 22.36 & \resulttok{8,594}
 & 55.56 & \resulttok{22,057} & 60.63 & \resulttok{17,463} \\
200 & 17.85 & \resulttok{3,428} & 22.50 & \resulttok{8,537}
 & 49.58 & \resulttok{24,701} & \bestacc{61.88} & \resulttok{17,265} \\
\bottomrule
\end{tabular*}
\end{table}

SRCL without DCE collapses to short, mostly incorrect responses: Qwen3-8B reaches at most
2.36\% with 1,116 tokens, and Qwen3-4B reaches 0.76\% with 531 tokens. SRCL therefore acts as a
concision objective only when paired with dynamic guidance.

\FloatBarrier

\section{Teacher-Update Schedules}
\label{app:teacher-trajectories}

We isolate the effect of teacher refresh by comparing frozen, exponential-moving-average (EMA),
periodic-snapshot, and fully dynamic teachers under the same Qwen3-8B setup. For EMA, the teacher
after optimizer step $t$ is
\begin{equation}
 \theta_T^{(t)}
 =d\,\theta_T^{(t-1)}+(1-d)\,\theta_S^{(t)},
 \label{eq:ema-teacher}
\end{equation}
where a smaller decay $d$ follows the student more closely. Periodic refresh instead copies the
current student into the teacher every $m$ optimizer steps and keeps the teacher fixed between
updates. The frozen teacher is never refreshed, while the dynamic teacher is synchronized at every
step. \Cref{tab:teacher-best} summarizes task-level performance, while
\cref{tab:ema-trajectories,tab:periodic-trajectories} give the complete Average@12 trajectories.

\begin{table}[H]
\centering
\caption{Task-level teacher-refresh comparison. Results are shown at the listed checkpoint;
Qwen3-8B includes frozen, EMA, and fully dynamic teachers, while Qwen3-4B compares frozen and
fully dynamic updates. Accuracy (Acc., \%) and mean generated tokens (Tok.) are reported separately.}
\label{tab:teacher-best}
\footnotesize
\setlength{\tabcolsep}{1.5pt}
\renewcommand{\arraystretch}{1.06}
\begin{tabular*}{0.99\textwidth}{@{\extracolsep{\fill}}lr*{5}{r@{\hspace{1.6pt}}r}@{}}
\toprule
\textbf{Teacher} & \textbf{Step}
 & \multicolumn{2}{c}{\textbf{AIME24}} & \multicolumn{2}{c}{\textbf{AIME25}}
 & \multicolumn{2}{c}{\textbf{AIME26}} & \multicolumn{2}{c}{\textbf{HMMT25}}
 & \multicolumn{2}{c}{\textcolor{metablue}{\textbf{Average}}} \\
\cmidrule(lr){3-4}\cmidrule(lr){5-6}\cmidrule(lr){7-8}\cmidrule(lr){9-10}\cmidrule(l){11-12}
 & & Acc. & Tok. & Acc. & Tok. & Acc. & Tok. & Acc. & Tok.
 & \textcolor{metablue}{Acc.} & \textcolor{metablue}{Tok.} \\
\midrule
\rowcolor{metablue!7}
\multicolumn{12}{@{}l}{\textcolor{metablue}{\sffamily\bfseries Qwen3--8B}} \\
Frozen $\theta_0$ & 50
 & 53.06 & \resulttok{8,600} & 42.22 & \resulttok{8,538}
 & 44.72 & \resulttok{8,899} & 24.17 & \resulttok{8,856}
 & 41.04 & \resulttok{8,723} \\
EMA $d=0.3$ & 150
 & 73.33 & \resulttok{17,984} & 69.72 & \resulttok{19,365}
 & 69.17 & \resulttok{18,341} & \bestacc{49.17} & \resulttok{23,276}
 & 65.35 & \resulttok{19,741} \\
EMA $d=0.5$ & 100
 & 71.11 & \resulttok{18,634} & 68.06 & \resulttok{19,921}
 & \bestacc{72.50} & \resulttok{18,647} & 46.39 & \resulttok{23,359}
 & 64.51 & \resulttok{20,140} \\
EMA $d=0.7$ & 100
 & \bestacc{73.89} & \resulttok{16,264} & 67.22 & \resulttok{18,567}
 & 70.83 & \resulttok{17,089} & 48.06 & \resulttok{21,728}
 & 65.00 & \resulttok{18,412} \\
\rowcolor{metateal!4}
Dynamic $\theta_k$ & 100
 & \bestacc{73.89} & \resulttok{15,640} & \bestacc{71.94} & \resulttok{17,241}
 & 71.94 & \resulttok{16,158} & 46.11 & \resulttok{21,205}
 & \bestacc{65.97} & \resulttok{17,561} \\
\midrule
\rowcolor{metablue!7}
\multicolumn{12}{@{}l}{\textcolor{metablue}{\sffamily\bfseries Qwen3--4B}} \\
Frozen $\theta_0$ & 30
 & 36.67 & \resulttok{7,496} & 27.22 & \resulttok{5,869}
 & 26.67 & \resulttok{7,194} & 18.61 & \resulttok{4,867}
 & 27.29 & \resulttok{6,357} \\
\rowcolor{metateal!4}
Evolving $\theta_k$ & 200
 & \bestacc{74.17} & \resulttok{15,273} & \bestacc{59.44} & \resulttok{17,466}
 & \bestacc{69.72} & \resulttok{16,181} & \bestacc{44.17} & \resulttok{20,143}
 & \bestacc{61.88} & \resulttok{17,265} \\
\bottomrule
\end{tabular*}
\end{table}

\begin{table}[H]
\centering
\caption{Qwen3-8B trajectories with exponential-moving-average teachers. Each pair reports
four-benchmark Average@12 accuracy (Acc., \%) and mean generated tokens (Tok.).}
\label{tab:ema-trajectories}
\small
\setlength{\tabcolsep}{4.5pt}
\renewcommand{\arraystretch}{1.08}
\begin{tabular*}{0.86\textwidth}{@{\extracolsep{\fill}}r*{3}{rr}@{}}
\toprule
\textbf{Step}
 & \multicolumn{2}{c}{\textbf{EMA $d=0.3$}}
 & \multicolumn{2}{c}{\textbf{EMA $d=0.5$}}
 & \multicolumn{2}{c}{\textbf{EMA $d=0.7$}} \\
\cmidrule(lr){2-3}\cmidrule(lr){4-5}\cmidrule(l){6-7}
 & Acc. & Tok. & Acc. & Tok. & Acc. & Tok. \\
\midrule
10  & 23.06 & \resulttok{4,916}  & 24.31 & \resulttok{4,772}  & 22.50 & \resulttok{4,603} \\
20  & 32.71 & \resulttok{6,972}  & 31.18 & \resulttok{7,202}  & 31.88 & \resulttok{7,051} \\
30  & 45.69 & \resulttok{11,164} & 46.53 & \resulttok{10,590} & 41.67 & \resulttok{9,589} \\
50  & 62.78 & \resulttok{18,760} & 63.33 & \resulttok{18,890} & 60.69 & \resulttok{16,613} \\
100 & 64.17 & \resulttok{19,729} & \bestacc{64.51} & \resulttok{20,140}
    & \bestacc{65.00} & \resulttok{18,412} \\
150 & \bestacc{65.35} & \resulttok{19,741} & 62.29 & \resulttok{21,161}
    & 62.99 & \resulttok{18,766} \\
200 & 63.75 & \resulttok{20,002} & 64.24 & \resulttok{21,314}
    & 62.71 & \resulttok{18,597} \\
\bottomrule
\end{tabular*}
\end{table}

\begin{table}[H]
\centering
\caption{Qwen3-8B trajectories with periodic hard-snapshot teachers. Interval $m$ copies the
current student into the teacher every $m$ optimizer steps and keeps it fixed between updates.}
\label{tab:periodic-trajectories}
\small
\setlength{\tabcolsep}{4.5pt}
\renewcommand{\arraystretch}{1.08}
\begin{tabular*}{0.86\textwidth}{@{\extracolsep{\fill}}r*{3}{rr}@{}}
\toprule
\textbf{Step}
 & \multicolumn{2}{c}{\textbf{Interval 10}}
 & \multicolumn{2}{c}{\textbf{Interval 20}}
 & \multicolumn{2}{c}{\textbf{Interval 30}} \\
\cmidrule(lr){2-3}\cmidrule(lr){4-5}\cmidrule(l){6-7}
 & Acc. & Tok. & Acc. & Tok. & Acc. & Tok. \\
\midrule
10  & 24.17 & \resulttok{4,609}  & 22.99 & \resulttok{4,802}  & 23.40 & \resulttok{4,747} \\
20  & 31.60 & \resulttok{6,864}  & 31.94 & \resulttok{6,923}  & 31.39 & \resulttok{6,475} \\
30  & 43.54 & \resulttok{10,023} & 42.57 & \resulttok{9,342}  & 41.39 & \resulttok{8,931} \\
50  & 60.76 & \resulttok{16,031} & 57.57 & \resulttok{15,867} & 49.17 & \resulttok{11,195} \\
100 & \bestacc{63.47} & \resulttok{18,355} & 62.43 & \resulttok{17,597}
    & 56.46 & \resulttok{13,178} \\
150 & 62.50 & \resulttok{19,469} & 63.13 & \resulttok{16,878}
    & 59.44 & \resulttok{15,582} \\
200 & 62.36 & \resulttok{19,499} & \bestacc{63.26} & \resulttok{16,856}
    & \bestacc{60.00} & \resulttok{16,026} \\
\bottomrule
\end{tabular*}
\end{table}

Every refresh strategy substantially improves over the frozen teacher. EMA $d=0.3$ reaches the
highest EMA accuracy at 65.35\%, while $d=0.7$ offers the better accuracy--length balance at
65.00\% with 18,412 tokens. Periodic snapshots are weaker, reaching at most 63.47\%. Updating the
teacher every step performs best overall at 65.97\% with 17,561 tokens. Because each schedule is
represented by one training trajectory, small differences should not be overinterpreted.

\begin{figure}[H]
    \centering
    \includegraphics[width=0.55\textwidth]{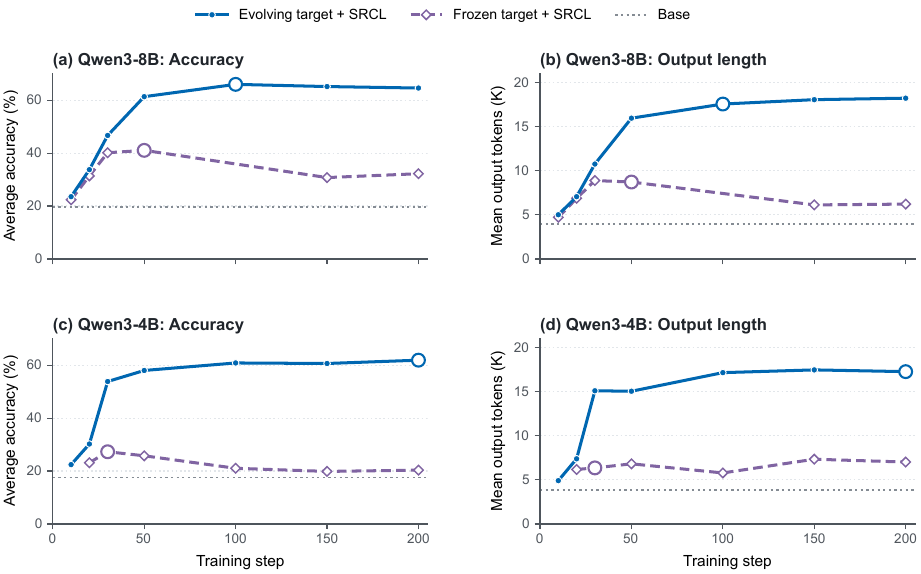}
    \caption{Frozen and fully dynamic DCE+SRCL trajectories at 8B (top) and 4B (bottom). Solid
    blue refreshes the privileged teacher every step, dashed purple keeps it frozen, and the dotted
    line denotes the base model.}
    \label{fig:teacher-ablation}
\end{figure}

\FloatBarrier

\section{Effect of Privileged-Context Placement}
\label{app:context-serialization}

The placement ablation separates teacher refresh from the representation of the verified solution.
\Cref{tab:context-teacher-placement} reports the complete Qwen3-8B comparison across the three
prompt formats illustrated in \cref{fig:prompt-serialization}. Every dynamic variant outperforms
its frozen counterpart. With a dynamic teacher, assistant-side prefill is strongest,
reference-last is intermediate, and instruction-last is weakest. Under a frozen teacher,
reference-last performs best but remains far below the dynamic variants.

\begin{table}[H]
\centering
\caption{Teacher refresh and verified-solution placement on Qwen3-8B. Each cell reports
Average@12 accuracy (\%) / mean generated tokens under non-thinking, 32K evaluation. The three
prompt formats are illustrated in \cref{fig:prompt-serialization}.}
\label{tab:context-teacher-placement}
\scriptsize
\setlength{\tabcolsep}{2.0pt}
\renewcommand{\arraystretch}{1.10}
\begin{tabular*}{\textwidth}{@{\extracolsep{\fill}}llccccc@{}}
\toprule
\textbf{Teacher} & \textbf{GOLD placement} & \textbf{AIME24} & \textbf{AIME25}
& \textbf{AIME26} & \textbf{HMMT25} & \textcolor{metablue}{\textbf{Average}} \\
\midrule
\rowcolor{metateal!5}
\textbf{Dynamic} & \textbf{Assistant-side}
& \bestacc{73.89} / 15,640 & \bestacc{71.94} / 17,241
& \bestacc{71.94} / 16,158 & \bestacc{46.11} / 21,205
& \bestacc{65.97} / 17,561 \\
Dynamic & Reference-last
& 71.94 / 15,685 & 67.50 / 17,706 & 66.94 / 16,737 & 40.56 / 21,645
& 61.74 / 17,944 \\
Dynamic & Instruction-last
& 66.11 / 16,982 & 61.94 / 18,423 & 62.50 / 16,336 & 34.17 / 21,625
& 56.18 / 18,341 \\
\midrule
Frozen & Assistant-side
& 53.06 / 8,600 & 42.22 / 8,538 & 44.72 / 8,899 & 24.17 / 8,856
& 41.04 / 8,723 \\
Frozen & Reference-last
& 57.22 / 9,512 & 43.33 / 9,319 & 50.28 / 10,087 & 26.11 / 10,119
& 44.24 / 9,759 \\
Frozen & Instruction-last
& 45.83 / 7,394 & 30.83 / 7,015 & 33.89 / 7,658 & 19.17 / 6,223
& 32.43 / 7,072 \\
\bottomrule
\end{tabular*}
\end{table}

After generating $Y_0$ without $g$, we hold it fixed. At position $t$, both branches score the same
$Y_{0,<t}$; only the privileged branch also receives $g$, and neither observes future tokens.

\begin{table}[H]
\centering
\caption{Position-resolved signal on a common set of fixed, correct Qwen3-8B responses. Each cell is
$D_{\mathrm{KL}}(q_T\|p_S)/[-\log q_T(Y_{0,t})]$. KL measures teacher--student distribution shift;
NLL measures teacher support for the observed token. Larger KL means more shift, while lower NLL
means more support, not greater correctness. Each serialization column comes from its own trained
model rather than from rescoring one model under three prompt formats.}
\label{tab:context-signal}
\footnotesize
\setlength{\tabcolsep}{5pt}
\renewcommand{\arraystretch}{1.08}
\begin{tabular*}{0.96\textwidth}{@{\extracolsep{\fill}}lccc@{}}
\toprule
\textbf{Position in $Y_0$}
 & \textbf{Assistant-side}
 & \textbf{Reference-last}
 & \textbf{Instruction-last} \\
\midrule
1--32       & 0.8427 / 0.6425 & 0.4787 / 0.5150 & 0.4648 / 0.4857 \\
33--128     & 0.4092 / 0.4324 & 0.4177 / 0.4483 & 0.4341 / 0.4382 \\
129--512    & 0.2667 / 0.2806 & 0.4198 / 0.3957 & 0.4277 / 0.3849 \\
513--1024   & 0.1724 / 0.1945 & 0.2761 / 0.2638 & 0.2863 / 0.2610 \\
\bottomrule
\end{tabular*}
\end{table}

\Cref{tab:context-signal} shows a front-loaded, not uniformly stronger, intervention. At positions
1--32, assistant-side has the highest KL/NLL, indicating the largest distribution shift and less
probability on the observed $Y_0$ token. After position 32, both metrics are lowest, indicating
agreement with the fixed continuation rather than correctness. This pattern is consistent with
route initialization, while the frozen reversal in \cref{tab:context-teacher-placement} shows that
placement alone is insufficient.

\paragraph{Evolution of the privileged signal.}
We next track the same diagnostic over training. At each checkpoint, KL compares the
verified-solution-conditioned teacher distribution $q_T$ with the no-GOLD student distribution
$p_S$ on the same fixed correct trajectories. KL records how strongly the privileged context changes
the prediction, but not whether that change favors the observed correct continuation. We therefore
also report
\begin{equation}
    \Delta\mathrm{NLL}
    = \mathrm{NLL}_{\text{no-GOLD}}-\mathrm{NLL}_{\text{with-GOLD}},
    \label{eq:context-delta-nll}
\end{equation}
where a positive value means that privileged conditioning assigns greater likelihood to the fixed
correct continuation.

\begin{table}[H]
\centering
\caption{Training dynamics of privileged conditioning on Qwen3-8B. KL measures the magnitude of
the teacher's intervention, whereas $\Delta$NLL measures whether that intervention increases
likelihood on the fixed correct continuation. Each cell reports KL / $\Delta$NLL. Each
serialization is a separately trained model, so the curves provide training-dynamics evidence
rather than an isolated causal effect of prompt position.}
\label{tab:context-signal-dynamics}
\footnotesize
\setlength{\tabcolsep}{5pt}
\renewcommand{\arraystretch}{1.08}
\begin{tabular*}{0.94\textwidth}{@{\extracolsep{\fill}}rccc@{}}
\toprule
\textbf{Step} & \textbf{Assistant-side} & \textbf{Reference-last} & \textbf{Instruction-last} \\
\midrule
10  & 0.1303 / $+0.0414$ & 0.0655 / $-0.0010$ & 0.0663 / $-0.0039$ \\
20  & 0.1888 / $+0.1364$ & 0.1005 / $+0.0391$ & 0.1023 / $+0.0323$ \\
30  & 0.2556 / $+0.2474$ & 0.1495 / $+0.0713$ & 0.1466 / $+0.0506$ \\
50  & 0.3646 / $+0.3729$ & 0.2387 / $+0.2209$ & 0.2178 / $+0.1492$ \\
100 & 0.4038 / $+0.4139$ & 0.2207 / $+0.0266$ & 0.1947 / $+0.0026$ \\
150 & 0.4367 / $+0.6053$ & 0.1973 / $-0.0358$ & 0.2151 / $+0.1201$ \\
200 & 0.4371 / $+0.6474$ & 0.2140 / $-0.0228$ & 0.2122 / $+0.0990$ \\
\bottomrule
\end{tabular*}
\end{table}

The two metrics separate intervention strength from direction: KL asks how much GOLD changes the
prediction, while $\Delta$NLL asks whether that change favors the recorded correct continuation.
For assistant-side conditioning, both signals increase throughout training: KL rises monotonically
from 0.1303 to 0.4371, and $\Delta$NLL rises from $+0.0414$ to $+0.6474$. Assistant-side also has
the largest $\Delta$NLL at every checkpoint, with its advantage widening later in training. The two
user-side runs strengthen through step 50 but then saturate or regress. From step 50 to 100,
$\Delta$NLL increases from $+0.3729$ to $+0.4139$ for assistant-side, but falls from $+0.2209$ to
$+0.0266$ for reference-last and from $+0.1492$ to $+0.0026$ for instruction-last. At step 200,
the assistant-side improvement exceeds reference-last and instruction-last by 0.6702 and 0.5484,
respectively. Thus, the user-side contexts can continue to alter the output distribution without
reliably increasing likelihood on the recorded correct path.

These curves show three training trajectories progressively separating; they do not by themselves
identify prompt position as the cause because each column comes from a separately trained model.
Nor is larger KL synonymous with higher benchmark accuracy: assistant-side accuracy peaks before
its KL does. Finally, $\Delta$NLL measures alignment with the fixed correct trajectories used in
this probe, not final-answer accuracy or support for every valid derivation.

\FloatBarrier

\section{Qwen3-1.7B Training Stability}
\label{app:qwen17}

At 1.7B, both recursive variants improve rapidly but become unstable later in training: DCE peaks
at step 30, while DCE+SRCL peaks at step 50 before declining. \Cref{tab:qwen17-appendix} reports
the complete trajectory and highlights the importance of early stopping at this scale.

\begin{table}[H]
\centering
\caption{Qwen3-1.7B checkpoint trajectories under non-thinking, 32K, Average@12 evaluation.
Adjacent columns report accuracy (Acc., \%) and mean generated tokens (Tok.); blue bold marks the
highest accuracy for each method.}
\label{tab:qwen17-appendix}
\footnotesize
\setlength{\tabcolsep}{3pt}
\renewcommand{\arraystretch}{0.94}
\begin{tabular*}{0.96\textwidth}{@{\extracolsep{\fill}}l*{4}{r@{\hspace{2.5pt}}r}@{}}
\toprule
\textbf{Step} & \multicolumn{2}{c}{\textbf{OPSD}} & \multicolumn{2}{c}{\textbf{GRPO}}
 & \multicolumn{2}{c}{\textcolor{metablue}{\textbf{DCE}}}
 & \multicolumn{2}{c}{\textcolor{metateal}{\textbf{+ SRCL}}} \\
\cmidrule(lr){2-3}\cmidrule(lr){4-5}\cmidrule(lr){6-7}\cmidrule(l){8-9}
 & Acc. & Tok. & Acc. & Tok. & Acc. & Tok. & Acc. & Tok. \\
\midrule
10  & 9.72 & \resulttok{3,359} & 9.10 & \resulttok{3,077}
 & 11.46 & \resulttok{4,099} & 12.22 & \resulttok{4,266} \\
20  & 9.79 & \resulttok{4,025} & 9.31 & \resulttok{3,122}
 & 15.76 & \resulttok{6,159} & 16.46 & \resulttok{6,874} \\
30  & 9.38 & \resulttok{4,237} & 8.89 & \resulttok{3,095}
 & \bestacc{23.33} & \resulttok{12,600} & 26.32 & \resulttok{14,797} \\
50  & \bestacc{10.35} & \resulttok{5,380} & 9.03 & \resulttok{3,151}
 & 22.57 & \resulttok{18,930} & \bestacc{26.88} & \resulttok{19,504} \\
100 & 9.58 & \resulttok{5,749} & \bestacc{9.58} & \resulttok{2,959}
 & 1.53 & \resulttok{30,241} & 25.90 & \resulttok{23,412} \\
150 & 8.47 & \resulttok{6,222} & 8.82 & \resulttok{2,916}
 & 21.94 & \resulttok{29,370} & 20.35 & \resulttok{24,059} \\
200 & 9.03 & \resulttok{6,282} & 8.89 & \resulttok{2,816}
 & 16.25 & \resulttok{31,033} & 1.53 & \resulttok{32,005} \\
\bottomrule
\end{tabular*}
\end{table}

\FloatBarrier

\section{Fixed-Trace Probe Values}
\label{app:eos-probe-values}

\Cref{fig:eos-probe-curves} summarizes endpoint EOS and reflection-cue probabilities on fixed
cohorts from AIME24, AIME25, and AIME26. The cohorts contain 227, 212, and 245 incorrect responses
and 879, 805, and 583 observed revision events, respectively. At each event, $r_t$ is the token
that appears when the stored trajectory begins to revise, such as \texttt{Wait}.
\Cref{tab:eos-probe-values} reports the complete AIME26 values, including EOS probability at the
same pre-reflection positions.

\begin{table}[H]
\centering
\caption{AIME26 next-token probabilities on 245 fixed incorrect responses containing 583 observed
reflection events. Student and teacher probabilities are reported separately; pre-reflection EOS
values are shown in units of $10^{-12}$.}
\label{tab:eos-probe-values}
\footnotesize
\setlength{\tabcolsep}{5.5pt}
\renewcommand{\arraystretch}{1.03}
\begin{tabular*}{0.96\textwidth}{@{\extracolsep{\fill}}lrrrrrr@{}}
\toprule
\textbf{Checkpoint}
 & \multicolumn{2}{c}{\textbf{Wrong-response endpoint}}
 & \multicolumn{4}{c}{\textbf{Before observed reflection cue}} \\
\cmidrule(lr){2-3}\cmidrule(l){4-7}
 & \multicolumn{2}{c}{$p(\eos)$ (\%)}
 & \multicolumn{2}{c}{$p(r_t)$ (\%)}
 & \multicolumn{2}{c}{$p(\eos)$ ($10^{-12}$)} \\
\cmidrule(lr){2-3}\cmidrule(lr){4-5}\cmidrule(l){6-7}
 & Student & Teacher & Student & Teacher & Student & Teacher \\
\midrule
Base & 88.71 & 85.57 & 29.12 & 37.22 & 26.5  & 35.3 \\
s10  & 86.28 & 83.44 & 41.11 & 48.00 & 10.3  & 15.8 \\
s20  & 80.79 & 79.01 & 59.98 & 64.04 & 1.14  & 2.67 \\
s30  & 68.73 & 68.72 & 73.83 & 75.93 & 0.114 & 0.378 \\
s50  & 54.52 & 57.40 & 80.99 & 81.33 & 0.299 & 0.285 \\
s100 & 33.10 & 37.24 & 81.13 & 81.35 & 0.984 & 0.985 \\
s150 & 26.28 & 35.81 & 80.39 & 81.02 & 0.866 & 1.02 \\
s200 & 22.75 & 34.87 & 79.93 & 80.84 & 0.589 & 1.04 \\
\bottomrule
\end{tabular*}
\end{table}

\FloatBarrier

\section{EOS-Penalty Ablation}
\label{app:eos-penalty}

\paragraph{Objective.}
\label{sec:late-eos-method}
To test whether concision can be induced directly, we add an EOS-specific auxiliary loss rather
than learning from SRCL rewrites. A verified response is truncated after its final balanced boxed
expression and retained as $y_i^{\mathrm{cut}}$ only if the answer judge still accepts it. Each
eligible response receives one EOS target:
\begin{equation}
 \mathcal{L}_{\mathrm{EOS}}^{(k)}
 =\begin{cases}
 \dfrac{1}{N_{\mathrm{valid}}}\displaystyle\sum_{i=1}^{N_{\mathrm{valid}}}
 \left[1-p_{\theta_k}\!\left(\eos\mid x_i,y_i^{\mathrm{cut}}\right)\right],
 & N_{\mathrm{valid}}>0,\\[6pt]
 0, & N_{\mathrm{valid}}=0.
 \end{cases}
 \label{eq:late-eos}
\end{equation}
The loss is averaged over eligible responses and set to zero when none are available. We evaluate
it as an alternative to SRCL rather than as part of the main method.

\eospenaltyresults

\FloatBarrier

\section{Cross-Scale Evaluation of OPSD-TTS}
\label{app:tts-results}

\Cref{tab:opsd-tts-full} reports OPSD-TTS at exact 8K and 16K output budgets. Doubling the forced
budget changes Average@12 by only 0.07, 1.05, and 2.29 points at 1.7B, 4B, and 8B, respectively,
and remains far below DCE. Additional generation alone therefore does not reproduce the benefit of
co-evolving training.

\begin{table}[H]
\centering
\caption{Task-level OPSD-TTS results at exact 8K and 16K output budgets. Accuracy (Acc., \%) and
generated tokens (Tok.) are reported separately; blue bold marks the higher Average@12 accuracy.}
\label{tab:opsd-tts-full}
\footnotesize
\setlength{\tabcolsep}{1.1pt}
\renewcommand{\arraystretch}{0.92}
\begin{tabular*}{0.99\textwidth}{@{\extracolsep{\fill}}ll*{5}{r@{\hspace{1.1pt}}r}@{}}
\toprule
\textbf{Model} & \textbf{Budget}
 & \multicolumn{2}{c}{\textbf{AIME24}} & \multicolumn{2}{c}{\textbf{AIME25}}
 & \multicolumn{2}{c}{\textbf{AIME26}} & \multicolumn{2}{c}{\textbf{HMMT25}}
 & \multicolumn{2}{c}{\textcolor{metablue}{\textbf{Average}}} \\
\cmidrule(lr){3-4}\cmidrule(lr){5-6}\cmidrule(lr){7-8}\cmidrule(lr){9-10}\cmidrule(l){11-12}
 & & Acc. & Tok. & Acc. & Tok. & Acc. & Tok. & Acc. & Tok.
 & \textcolor{metablue}{Acc.} & \textcolor{metablue}{Tok.} \\
\midrule
\rowcolor{metablue!5}
1.7B & 8K
 & 14.17 & \resulttok{8,192} & 8.06 & \resulttok{8,192}
 & 8.06 & \resulttok{8,192} & 6.39 & \resulttok{8,192}
 & 9.17 & \resulttok{8,192} \\
1.7B & 16K
 & 14.72 & \resulttok{16,384} & 7.50 & \resulttok{16,384}
 & 8.89 & \resulttok{16,384} & 5.83 & \resulttok{16,384}
 & \bestacc{9.24} & \resulttok{16,384} \\
\midrule
\rowcolor{metablue!5}
4B & 8K
 & 28.06 & \resulttok{8,192} & 22.50 & \resulttok{8,192}
 & 23.61 & \resulttok{8,192} & 13.61 & \resulttok{8,192}
 & 21.94 & \resulttok{8,192} \\
4B & 16K
 & 30.28 & \resulttok{16,384} & 22.78 & \resulttok{16,384}
 & 25.56 & \resulttok{16,384} & 13.33 & \resulttok{16,384}
 & \bestacc{22.99} & \resulttok{16,384} \\
\midrule
\rowcolor{metablue!5}
8B & 8K
 & 43.06 & \resulttok{8,192} & 27.50 & \resulttok{8,192}
 & 28.33 & \resulttok{8,192} & 15.00 & \resulttok{8,192}
 & 28.47 & \resulttok{8,192} \\
8B & 16K
 & 46.39 & \resulttok{16,384} & 28.89 & \resulttok{16,384}
 & 31.11 & \resulttok{16,384} & 16.67 & \resulttok{16,384}
 & \bestacc{30.76} & \resulttok{16,384} \\
\bottomrule
\end{tabular*}
\end{table}

\FloatBarrier

\section{Reproducibility Details}
\label{app:reproducibility}

\Cref{tab:configuration} summarizes the data, optimization, generation, and evaluation settings
used throughout the main experiments; each ablation states its deviations explicitly.

\begingroup
\footnotesize
\setlength{\tabcolsep}{5pt}
\renewcommand{\arraystretch}{1.02}
\setlength{\LTleft}{\fill}
\setlength{\LTright}{\fill}
\begin{longtable}{@{}>{\raggedright\arraybackslash}p{0.19\textwidth}>{\raggedright\arraybackslash}p{0.75\textwidth}@{}}
\caption{Training and evaluation configuration used in the main experiments.}
\label{tab:configuration} \\
\toprule
\textbf{Component} & \textbf{Configuration} \\
\midrule
\endfirsthead
\multicolumn{2}{@{}l}{\textbf{Table \thetable}\quad Training and evaluation configuration
(continued).} \\
\toprule
\textbf{Component} & \textbf{Configuration} \\
\midrule
\endhead
\midrule
\multicolumn{2}{r@{}}{\footnotesize Continued on next page} \\
\endfoot
\bottomrule
\endlastfoot
\rowcolor{metablue!9}
\multicolumn{2}{@{}l}{\textcolor{metablue}{\sffamily\bfseries Models and data}} \\
Model family & Qwen3-1.7B, Qwen3-4B, Qwen3-8B, and Qwen3-14B; cross-family transfer uses
Gemma-4-12B-IT (\cref{app:gemma12b-details}). Student and privileged branches operate in
non-thinking mode. \\
Training data & The 14,717 mathematical-reasoning problems from OpenThoughts used by OPSD. DCE
constructs student and privileged views of each problem; these are two contexts for the same
training example, not separate problems. \\
Baseline setup & SFT and GRPO use the same training problems and boxed-answer prompt;
both run in non-thinking mode for 200 optimizer steps, saving at steps
$\{10,20,30,50,100,150,200\}$. They use global batch size 16 across eight GPUs and rank-128,
scale-256 LoRA with dropout 0.05 on attention and MLP projections; seed 42. \\
\rowcolor{metablue!9}
\multicolumn{2}{@{}l}{\textcolor{metablue}{\sffamily\bfseries Training}} \\
SFT baseline & Cross-entropy on the complete gold solution followed by EOS, with prompt tokens
masked; 12K target cap and 16,384-token sequence limit. We sweep learning rates
$\{5\!\times\!10^{-7},10^{-6},2\!\times\!10^{-6}\}$, together with the historical
$5\!\times\!10^{-6}$ run; no teacher, KD loss, or rollout engine is used. \\
GRPO baseline & Eight generations per problem, two policy iterations, temperature 1.2, and a 12K
maximum training-generation length. Training uses a binary final-answer reward, group
normalization, $\beta=0$, and learning rates $\{10^{-6},2\!\times\!10^{-6},5\!\times\!10^{-6}\}$.
Rollouts use colocated vLLM. \\
Tuned baseline settings & SFT uses $(2\!\times\!10^{-6},s20)$ at 8B,
$(5\!\times\!10^{-7},s20)$ at 4B, and $(5\!\times\!10^{-7},s30)$ at 1.7B; GRPO uses
$(2\!\times\!10^{-6},s100)$ at 4B and $(2\!\times\!10^{-6},s20)$ at 1.7B. \\
Privileged context & Current model weights; problem, first 512 reference tokens, transition, and the
retained on-policy response prefix. Gradients are stopped through the privileged branch. \\
Guidance objective & Forward KL for the main experiments; Reverse KL and JSD are evaluated as
ablations. \\
Qwen3 DCE optimization & AdamW, bfloat16, learning rate $5\times10^{-6}$, gradient clipping at 0.1, batch size
16, and rank-128 LoRA with scale 256 on attention and MLP projections; eight H100 GPUs. \\
Rollout and guidance & One 12K rollout per example with $T=1.1$, top-$p=0.95$, and top-$k=20$;
the retained prefix receives the DCE guidance objective. \\
SRCL rewrite & Greedy 12K rewrite conditioned on the problem and original response, without the
reference solution. Accepted targets must terminate naturally, be shorter, remain self-contained,
and pass structural and answer-verification filters. \\
Qwen3 outer loss weights & Main $\lambda_G=5\times10^4$; $\lambda_S=12.5,25,35,$ and $25$ for 14B,
8B, 4B, and 1.7B, respectively. \\
Training horizon & Up to 200 optimizer steps with seed 42; checkpoints are evaluated at steps
$\{10,20,30,50,100,150,200\}$ when available. \\
\rowcolor{metablue!9}
\multicolumn{2}{@{}l}{\textcolor{metablue}{\sffamily\bfseries Evaluation}} \\
Main protocol & Four 30-problem benchmarks, 12 samples per problem (360 generations per dataset;
1,440 total), non-thinking decoding, 32K generation cap, $T=1.0$, top-$p=0.8$, top-$k=-1$, and
repetition penalty 1.0. Accuracy uses the verified final answer; token counts include generated
output only and exclude prompts. \\
TTS control & OPSD checkpoints evaluated at exact 8K or 16K output budgets; a \texttt{Wait} cue
continues responses that terminate early. Each benchmark contains 360 generations, and token counts
exclude prompts. \\
\end{longtable}
\endgroup

\FloatBarrier

\section{Additional SRCL-Weight Sensitivity}
\label{app:srcl-weight-4b}

\Cref{tab:srcl-coefficient-4b} extends the SRCL-weight sweep to Qwen3-4B.

\begin{table}[H]
\centering
\caption{Qwen3-4B SRCL-weight sensitivity under non-thinking, 32K, Average@12 evaluation.
Accuracy (Acc., \%) and mean generated tokens (Tok.) are reported separately; pale teal marks the
main setting.}
\label{tab:srcl-coefficient-4b}
\footnotesize
\setlength{\tabcolsep}{1.8pt}
\renewcommand{\arraystretch}{1.08}
\begin{tabular*}{0.99\textwidth}{@{\extracolsep{\fill}}l*{5}{r@{\hspace{2.2pt}}r}@{}}
\toprule
\textbf{$\lambda_S$}
 & \multicolumn{2}{c}{\textbf{AIME24}} & \multicolumn{2}{c}{\textbf{AIME25}}
 & \multicolumn{2}{c}{\textbf{AIME26}} & \multicolumn{2}{c}{\textbf{HMMT25}}
 & \multicolumn{2}{c}{\textcolor{metablue}{\textbf{Average}}} \\
\cmidrule(lr){2-3}\cmidrule(lr){4-5}\cmidrule(lr){6-7}\cmidrule(lr){8-9}\cmidrule(l){10-11}
 & Acc. & Tok. & Acc. & Tok. & Acc. & Tok. & Acc. & Tok.
 & \textcolor{metablue}{Acc.} & \textcolor{metablue}{Tok.} \\
\midrule
25 & 65.83 & \resulttok{18,888} & 58.33 & \resulttok{20,431}
 & 67.78 & \resulttok{18,758} & 43.33 & \resulttok{23,574} & 58.82 & \resulttok{20,413} \\
30 & 67.22 & \resulttok{18,254} & 56.94 & \resulttok{20,374}
 & 68.33 & \resulttok{18,132} & 39.72 & \resulttok{22,597} & 58.06 & \resulttok{19,839} \\
32.5 & 65.28 & \resulttok{18,990} & 58.33 & \resulttok{20,562}
 & 69.17 & \resulttok{18,513} & 42.22 & \resulttok{23,556} & 58.75 & \resulttok{20,405} \\
\rowcolor{metateal!5}
\textbf{35} & \bestacc{74.17} & \resulttok{15,273}
 & \bestacc{59.44} & \resulttok{17,466} & \bestacc{69.72} & \resulttok{16,181}
 & \bestacc{44.17} & \resulttok{20,143} & \bestacc{61.88} & \resulttok{17,265} \\
37.5 & 68.06 & \resulttok{17,344} & 57.22 & \resulttok{18,789}
 & 67.22 & \resulttok{17,437} & 38.89 & \resulttok{21,279} & 57.85 & \resulttok{18,712} \\
50 & 66.11 & \resulttok{16,556} & 57.78 & \resulttok{17,735}
 & 64.44 & \resulttok{16,318} & 39.17 & \resulttok{21,490} & 56.88 & \resulttok{18,025} \\
\bottomrule
\end{tabular*}
\end{table}

The sweep peaks at $\lambda_S=35$, reaching 61.88\% Average@12 with 17,265 tokens. Relative to
$\lambda_S=25$, it improves accuracy by 3.06 points while using 3,148 fewer tokens; larger weights
then reduce accuracy, showing that the balance between guidance and concise-target learning remains
important at 4B.

\FloatBarrier

\section{Results on Additional Benchmarks}
\label{app:general-benchmarks}

We additionally evaluate Qwen3-8B and Qwen3-4B under Average@12 on three benchmarks:
MATH-500 \citep{hendrycks2021math}, GPQA-Diamond \citep{rein2023gpqa}, and AMC 2023.

\begin{table}[H]
\centering
\caption{Average@12 results on three additional benchmarks. GPQA-D denotes GPQA-Diamond;
accuracy (Acc., \%) and mean generated tokens (Tok.) are reported separately.}
\label{tab:general-benchmarks}
\footnotesize
\setlength{\tabcolsep}{1.8pt}
\renewcommand{\arraystretch}{1.08}
\begin{tabular*}{0.86\textwidth}{@{\extracolsep{\fill}}l*{3}{r@{\hspace{2.5pt}}r}@{}}
\toprule
\textbf{Method}
 & \multicolumn{2}{c}{\textbf{MATH-500}}
 & \multicolumn{2}{c}{\textbf{GPQA-D}}
 & \multicolumn{2}{c}{\textbf{AMC23}} \\
\cmidrule(lr){2-3}\cmidrule(lr){4-5}\cmidrule(l){6-7}
 & Acc. & Tok. & Acc. & Tok. & Acc. & Tok. \\
\midrule
\rowcolor{metablue!7}
\multicolumn{7}{@{}l}{\textcolor{metablue}{\sffamily\bfseries Qwen3--8B}} \\
Base
 & 84.22 & \resulttok{1,041} & 48.61 & \resulttok{1,457}
 & 70.83 & \resulttok{1,998} \\
OPSD
 & 88.58 & \resulttok{1,340} & 50.29 & \resulttok{1,877}
 & 78.54 & \resulttok{2,633} \\
\rowcolor{metateal!5}
\textbf{DCE+SRCL}
 & \bestacc{88.95} & \resulttok{4,043} & \bestacc{58.29} & \resulttok{4,296}
 & \bestacc{95.42} & \resulttok{7,604} \\
\midrule
\rowcolor{metablue!7}
\multicolumn{7}{@{}l}{\textcolor{metablue}{\sffamily\bfseries Qwen3--4B}} \\
Base
 & 84.02 & \resulttok{969} & 42.59 & \resulttok{1,375}
 & 69.38 & \resulttok{1,660} \\
OPSD
 & 85.45 & \resulttok{1,553} & 38.97 & \resulttok{3,047}
 & 70.83 & \resulttok{3,433} \\
\rowcolor{metateal!5}
\textbf{DCE+SRCL}
 & \bestacc{87.85} & \resulttok{4,735} & \bestacc{47.47} & \resulttok{2,740}
 & \bestacc{95.83} & \resulttok{7,589} \\
\bottomrule
\end{tabular*}
\end{table}

DCE+SRCL improves over OPSD on all three benchmarks at both scales, with the largest gains on
AMC23. The improved accuracy generally accompanies longer outputs on these broader tasks, unlike
the concision gains observed on the main benchmark suite.

\FloatBarrier

\FloatBarrier

\section{SRCL Rewrite Filtering and Acceptance}
\label{app:srcl-acceptance}

SRCL learns only from model-generated rewrites that satisfy its concision and correctness criteria. We audit
all 3,200 candidates generated during the Qwen3-8B DCE+SRCL run in
\cref{tab:srcl-acceptance}. The acceptance rate remains stable between 70.63\% and 73.13\% across
four consecutive 50-step windows, with 2,296 rewrites (71.75\%) retained overall. Accepted targets
are 83.74\% shorter than their source rollouts, whose mean length is 1,997 tokens. Of the 904
rejected candidates, 826 fail a structural check, 76 retain explicit revision language, and two do
not terminate naturally. Rejection categories record the first failed gate. A rejected rewrite
contributes no SRCL loss, although its original example still receives DCE training.

\begin{table}[H]
\centering
\caption{SRCL rewrite acceptance across the Qwen3-8B training trajectory. Rejection columns record
the first failed gate; Reflection denotes explicit revision markers such as \texttt{Wait},
\texttt{Actually}, and \texttt{Let me reconsider}.}
\label{tab:srcl-acceptance}
\scriptsize
\setlength{\tabcolsep}{6pt}
\renewcommand{\arraystretch}{0.96}
\begin{tabular}{lrrrrrr}
\toprule
\textbf{Steps} & \textbf{Candidates} & \textbf{Accepted} & \textbf{Rate}
& \textbf{Structural} & \textbf{Reflection} & \textbf{Non-stop} \\
\midrule
1--50    & 800 & 585 & 73.13\% & 191 & 22 & 2 \\
51--100  & 800 & 576 & 72.00\% & 208 & 16 & 0 \\
101--150 & 800 & 565 & 70.63\% & 214 & 21 & 0 \\
151--200 & 800 & 570 & 71.25\% & 213 & 17 & 0 \\
\midrule
\textbf{All} & \textbf{3,200} & \textbf{2,296} & \textbf{71.75\%}
& \textbf{826} & \textbf{76} & \textbf{2} \\
\bottomrule
\end{tabular}
\end{table}

\paragraph{Filtering pipeline.}
A candidate is retained only if it passes all four gates below.
\begin{enumerate}
    \item \textbf{Termination and compression.} The rewrite must terminate naturally, be nonempty,
    remain within the 12K limit, and contain fewer tokens than its source rollout.
    \item \textbf{Reflection-free rewriting.} A case-insensitive scan rejects explicit reconsideration,
    correction, restart, or repeated-verification phrases, including \texttt{Wait},
    \texttt{Actually}, \texttt{Correction}, \texttt{double-check}, and \texttt{start over}.
    \item \textbf{Answer validity.} The rewrite must contain at least 32 tokens and a balanced
    boxed answer near the end. The final box must be correct, and no
    earlier box may contain an incorrect answer.
    \item \textbf{Self-containment and repetition.} The rewrite may not begin as a continuation
    fragment, refer to omitted material, or exceed the implementation's repeated-phrase and
    repeated-line thresholds.
\end{enumerate}

\paragraph{Endpoint verification.}
The verifier extracts the first balanced boxed answer after the
final \texttt{</think>} delimiter, when present, and compares it with the gold answer using
\texttt{math\_verify}. If parsing fails, it falls back to case-insensitive exact matching after
whitespace removal; ratio notation $a:b$ is normalized to $a/b$. A separate structural check
examines every boxed expression, requiring the final box to be correct and rejecting candidates
with an earlier incorrect box.

\FloatBarrier

\section{Prompt Templates}
\label{app:prompt-templates}

The logical prompt templates are shown below, with model-specific chat-control tokens omitted.
All privileged variants use the same problem $x$, verified solution $g$, transition text, and
on-policy response $Y_0$; only the message role and ordering change. SRCL uses a separate rewrite
request.

\begin{tcolorbox}[
  breakable,
  width=0.98\textwidth,
  colback=metagold!3,
  colframe=metagold!45,
  boxrule=0.45pt,
  arc=1.5pt,
  left=5pt,right=5pt,top=2pt,bottom=2pt,
  title={\sffamily\bfseries SRCL rewrite prompt},
  fonttitle=\small]
\scriptsize\ttfamily\raggedright
SOURCE ROLLOUT\par
\{full\_rollout\}\par
END SOURCE ROLLOUT\par\smallskip
PROBLEM\par
\{problem\}\par
END PROBLEM\par\smallskip
Rewrite the source rollout into the shortest direct, self-contained, correct solution to the
problem. Preserve only reasoning needed to derive the final answer. Remove every failed branch,
retry, repeated calculation, and reflection phrase such as Wait, reconsider, actually, correction,
or start over. Do not mention the source rollout or omitted text. Do not add analysis about
rewriting. End with exactly one final \textbackslash boxed\{...\} answer. Output only the clean
solution.
\end{tcolorbox}

\enlargethispage{0.75\baselineskip}
\begin{figure}[H]
\centering
\captionsetup{skip=3pt,font=footnotesize}
\begin{tcolorbox}[
  width=0.98\textwidth,
  colback=metablue!3,
  colframe=metablue!35,
  boxrule=0.45pt,
  arc=1.5pt,
  left=6pt,right=6pt,top=2pt,bottom=2pt,
  title={\sffamily\bfseries Student and inference context},
  fonttitle=\small]
\footnotesize\raggedright
\textsf{\bfseries USER}\quad \texttt{Problem: }$x$\par
\texttt{Please reason step by step, and put your final answer within
\textbackslash boxed\{\}.}\par
\textsf{\bfseries ASSISTANT}\quad $Y_{0,<t}\;\longrightarrow\;\text{predict }Y_{0,t}$
\end{tcolorbox}

\vspace{0pt}
\begin{tcolorbox}[
  width=0.98\textwidth,
  colback=metateal!4,
  colframe=metateal!45,
  boxrule=0.45pt,
  arc=1.5pt,
  left=6pt,right=6pt,top=2pt,bottom=2pt,
  title={\sffamily\bfseries Assistant-side},
  fonttitle=\small]
\footnotesize\raggedright
\textsf{\bfseries USER}\quad \texttt{Problem: }$x$\par
\texttt{Please reason step by step, and put your final answer within
\textbackslash boxed\{\}.}\par
\textsf{\bfseries ASSISTANT}\par
\texttt{Here is a reference solution: }$g$\par
\texttt{After understanding the reference solution, please try to solve this problem using your own approach below:}\par
$Y_{0,<t}\;\longrightarrow\;\text{predict }Y_{0,t}$
\end{tcolorbox}

\vspace{0pt}
\begin{tcolorbox}[
  width=0.98\textwidth,
  colback=metagold!4,
  colframe=metagold!48,
  boxrule=0.45pt,
  arc=1.5pt,
  left=6pt,right=6pt,top=2pt,bottom=2pt,
  title={\sffamily\bfseries User-side: reference-last},
  fonttitle=\small]
\footnotesize\raggedright
\textsf{\bfseries USER}\quad \texttt{Problem: }$x$\par
\texttt{Please reason step by step, and put your final answer within
\textbackslash boxed\{\}.}\par
\texttt{Here is a reference solution: }$g$\par
\texttt{After understanding the reference solution, please try to solve this problem using your own approach below:}\par
\textsf{\bfseries ASSISTANT}\quad $Y_{0,<t}\;\longrightarrow\;\text{predict }Y_{0,t}$
\end{tcolorbox}

\vspace{0pt}
\begin{tcolorbox}[
  width=0.98\textwidth,
  colback=metapurple!4,
  colframe=metapurple!45,
  boxrule=0.45pt,
  arc=1.5pt,
  left=6pt,right=6pt,top=2pt,bottom=2pt,
  title={\sffamily\bfseries User-side: instruction-last},
  fonttitle=\small]
\footnotesize\raggedright
\textsf{\bfseries USER}\quad \texttt{Problem: }$x$\par
\texttt{Here is a reference solution: }$g$\par
\texttt{After understanding the reference solution, please try to solve this problem using your own approach below:}\par
\texttt{Please reason step by step, and put your final answer within
\textbackslash boxed\{\}.}\par
\textsf{\bfseries ASSISTANT}\quad $Y_{0,<t}\;\longrightarrow\;\text{predict }Y_{0,t}$
\end{tcolorbox}

\caption{Prompt orderings for the placement ablation, with chat-control tokens omitted.
Assistant-side places $g$ in prior assistant context; the user-side variants place it before or
after the task instruction. Here $Y_{0,<t}$ is the fixed prefix used to score token $t$.}
\label{fig:prompt-serialization}
\end{figure}

\end{document}